\documentclass{ws-ijalp}

\usepackage{float}
\usepackage{tabularx}
\usepackage[utf8]{inputenc} 
\usepackage[T5]{fontenc}    
\usepackage{caption}
\usepackage{multicol}
\usepackage{subfigure}
\usepackage{appendix}
\usepackage{longtable}
\usepackage{booktabs}
\usepackage{color,soulutf8}
\usepackage{multirow}
\usepackage{graphicx}%
\usepackage{subcaption}
\usepackage{amsmath,amssymb,amsfonts}%
\usepackage{amsthm}%
\usepackage{mathrsfs}%
\usepackage{xcolor}%
\usepackage{textcomp}%
\usepackage{manyfoot}%
\usepackage{makecell}%
\usepackage{multicol}
\usepackage{supertabular}
\usepackage{lipsum}
\usepackage{soul}
\usepackage{academicons}
\usepackage{float}
\usepackage{pifont}
\usepackage{url}
\usepackage[verbose]{hyperref}

\begin{document}

\markboth{Luu et al.}{ViMMRC 2.0 - Enhancing Machine Reading Comprehension on Vietnamese Literature Text}


\title{ViMMRC 2.0 - Enhancing Machine Reading Comprehension on Vietnamese Literature Text}

\author{Son T. Luu\footnote{ORCID: \url{https://orcid.org/0000-0002-1231-5865}}}

\address{Faculty of Information Science and Engineering, University of Information Technology, Ho Chi Minh City, Vietnam \\ Vietnam National University, Ho Chi Minh City, Vietnam
\\
sonlt@uit.edu.vn}

\author{Khoi Trong Hoang}

\address{Faculty of Information Science and Engineering, University of Information Technology, Ho Chi Minh City, Vietnam \\ Vietnam National University, Ho Chi Minh City, Vietnam
\\
19521706@gm.uit.edu.vn}

\author{Tuong Quang Pham}

\address{Faculty of Information Science and Engineering, University of Information Technology, Ho Chi Minh City, Vietnam \\ Vietnam National University, Ho Chi Minh City, Vietnam
\\
19522499@gm.uit.edu.vn}

\author{Kiet Van Nguyen\footnote{ORCID: \url{https://orcid.org/0000-0002-8456-2742}}}

\address{Faculty of Information Science and Engineering, University of Information Technology, Ho Chi Minh City, Vietnam \\ Vietnam National University, Ho Chi Minh City, Vietnam
\\
kietnv@uit.edu.vn}

\author{Ngan Luu-Thuy Nguyen\footnote{Corresponding author. ORCID: \url{https://orcid.org/0000-0003-3931-849X}}}

\address{Faculty of Information Science and Engineering, University of Information Technology, Ho Chi Minh City, Vietnam \\ Vietnam National University, Ho Chi Minh City, Vietnam
\\
ngannlt@uit.edu.vn}

\maketitle


\begin{abstract}
Machine reading comprehension has been an interesting and challenging task in recent years, with the purpose of extracting useful information from texts. To attain the computer ability to understand the reading text and answer relevant information, we introduce ViMMRC 2.0 - an extension of the previous ViMMRC for the task of multiple-choice reading comprehension in Vietnamese Textbooks which contain the reading articles for students from Grade 1 to Grade 12. This dataset has 699 reading passages which are prose and poems, and 5,273 questions. The questions in the new dataset are not fixed with four options as in the previous version. Moreover, the difficulty of questions is increased, which challenges the models to find the correct choice. The computer must understand the whole context of the reading passage, the question, and the content of each choice to extract the right answers. Hence, we propose a multi-stage approach that combines the multi-step attention network (MAN) with the natural language inference (NLI) task to enhance the performance of the reading comprehension model. Then, we compare the proposed methodology with the baseline BERTology models on the new dataset and the ViMMRC 1.0. From the results of the error analysis, we found that the challenge of the reading comprehension models is understanding the implicit context in texts and linking them together in order to find the correct answers.  Finally, we hope our new dataset will motivate further research to enhance the ability of computers to understand the Vietnamese language.
\end{abstract}


\section{Introduction}
Machine reading comprehension (MRC) is the task that allows the computer to find the correct information based on the input question. This task can be categorized as a question-answering task. According to \cite{dzendzik-etal-2021-english}, MRC consists of three components: the passage, the question, and the answer. The question is the input query from users in natural language texts, the passage is the context, and the answer is the extracted information. Concerning the type of answers, the MRC task is classified as Extractive, Multiple-choices, Boolean, and Free form \cite{dzendzik-etal-2021-english}. Above all, multiple-choice reading comprehension is one of the earliest studied tasks, which began in 2013 with the appearance of the MCTest corpus \cite{richardson-etal-2013-mctest}. The answers in this task are given as a list of options, in which the content can be extractive, free-form, or even in yes/no form. The computer must choose at least one option as the correct one. The multiple-choice task aims to enhance the ability of computers to comprehend the passage and the question to give the correct choices. In Vietnamese, the authors in \cite{9247161} provided the ViMMRC dataset (we called ViMMRC 1.0 in our paper) to evaluate the ability to understand human texts and answer the questions as multiple choices. This is the first dataset for multiple-choice reading comprehension on Vietnamese texts. However, the ViMMRC 1.0 dataset is limited to the level of reading comprehension, in which the reading passages only focus on students in primary school (from Grade 1 to Grade 5). From the error analysis, the authors of the dataset mentioned that increasing the size of the dataset will improve the performance of the model \cite{9247161}. 

Following the previous work in \cite{9247161}, we introduce ViMMRC 2.0, a new corpus for multiple-choice reading comprehension of Vietnamese text in this paper. This dataset inherits from the ViMMRC 1.0 \cite{9247161}. The ViMMRC 2.0 dataset has 699 reading passages and 5,273 questions. The text from the dataset is collected from the literature textbook for Vietnamese students from $1^{st}$ to $12^{th}$ grades, published by Vietnam Education Publishing House (VEPH) under the Vietnam Minister of Education and Training (VN-MoET). In this new dataset, the reading texts, questions, and answers are for students from primary to high school. To gain an in-depth understanding of the questions, we divide the questions into four types: the Wh-question, the listed options question, the Yes/No question, and Others. The detailed type of the dataset is shown in Section \ref{dataset}. Based on the analysis of reading passages and questions, we show the challenge of the new ViMMRC 2.0 dataset compared to ViMMRC 1.0. 

To extract the correct choices, the machine must understand the whole context of the reading passages and link the relevant facts between the reading passages, the questions, and the content in the options. We propose a multi-stage approach based on previous work in \cite{jin2020mmm}, which uses the BERT \cite{devlin-etal-2019-bert} as a language model to understand the texts combined with the multi-step attention (MAN) mechanism and natural language inference (NLI) to capture the whole context of the reading texts and the questions. Finally, we take an in-depth analysis of the predictions to study the ability of the models on the reading comprehension task. 
 
In general, our main contributions to this paper are described below:
\begin{itemize}
\item We proposed ViMMRC 2.0 - an enhanced version of the previous dataset for Vietnamese multiple-choice reading comprehension with 699 passages and 5,273 questions. 
\item We analyze the reading passages and questions in our new dataset to explore the linguistic aspect of Vietnamese text.
\item We propose a multi-stage approach integrating the Multi-step attention network (MAN) and Natural language inference (NLI) to enhance the performance of the reading comprehension model. 
\item We implement our proposed method and compare it with the previous baseline models to have an in-depth understanding of the ability of the computer in machine reading comprehension.
\end{itemize}

The paper is structured as follows. Section \ref{related_work} takes an overview of previous works on Vietnamese MRC including datasets and methods for constructing MRC models. Section \ref{dataset} introduces the creation of the ViMMRC 2.0 dataset and the in-depth analysis of the reading passages and questions. Section \ref{method} described our methodology including the multi-stage paradigm to improve the performance of the MRC model. Section \ref{results} illustrates our empirical results of the proposed method on the ViMMRC 2.0 dataset. Next, Section \ref{error_analysis} analyzes the performance of the proposed MRC models based on the ViMMRC 2.0 dataset. Finally, Section \ref{conclusion} concludes our works and suggests future studies for the multiple-choice MRC task.
\section{Related Works}
\label{related_work}
\subsection{Datasets}

In Vietnamese, the ViMMRC \cite{9247161} and ViQuAD \cite{nguyen-etal-2020-vietnamese} are two early datasets for training the machine reading comprehension models. The ViQuAD \cite{nguyen-etal-2020-vietnamese} contains 23,000 question-answer pairs from 5,109 passages. From the ViQuAD, the answers are directly extracted from the reading passages based on the context of the questions. It is called extractive machine reading comprehension (or extractive MRC). However, the weakness of extractive MRC is the impossibility of finding the correct answers to questions that are not explicitly stated in the reading passage. These are called unanswerable questions. To solve that problem, the UIT-ViQuAD 2.0 dataset \cite{JCSCE} is constructed with the appearance of 9,217 unanswerable questions. This dataset has been released in a shared task named the VLSP 2021-ViMRC Challenge\footnote{\url{https://aihub.vn/competitions/35}}. 
In contrast, the significant challenge of the ViMMRC \cite{9247161} is the limited size of the dataset, which affects the performance of the reading comprehension ability of the computers. The ViMMRC dataset only contains 2,783 questions with 417 reading passages from student textbooks for Grades 1 to Grade 5. This is our motivation in this paper for constructing a new multiple-choice reading comprehension dataset with more reading passages from higher grades for students and more various questions to enhance the ability of computers for the task. 

\begin{table}[ht]
    \caption{Vietnamese MRC datasets}
    \label{tab_vi_dataset}
    \resizebox{\textwidth}{!}{
    \begin{tabular}{lp{2.5cm}p{4cm}p{4cm}}
        \toprule
        \textbf{Datasets} & \textbf{Types} & \textbf{Size} & \textbf{Source} \\
        \midrule
        ViMMRC \cite{9247161} & Multiple choice &  417 reading passages and 2,783 questions & Primary School Students Textbook  \\ \hline
        UIT-ViQuAD \cite{nguyen-etal-2020-vietnamese} & Extractive & 5,109 reading passages and 23,074 questions & Wikipedia (open domain) \\ \hline
        ViCoQA \cite{10.1007/978-3-030-88113-9_44} & Conversational MRC & 2,0000 conversations and 10,0000 questions & Health News \\ \hline
        UIT-ViQuAD 2.0 \cite{JCSCE} & Extractive & 5,173 passages and 35,990 questions (including 9,217 unanswerable questions) & Wikipedia (open domain) \\ \hline
        ViHealthQA \cite{10.1007/978-3-031-10986-7_30} & Retrieval & 10,015 question-answer passage pairs & Health articles from news websites\\ \hline
        UIT-ViWikiQA \cite{10.1007/978-3-030-82147-0_42} & Extractive (sentence) & 5.109 passage and 3.074 question-answers & Wikipedia (open domain)\\ \hline
        UIT-ViNewsQA \cite{10.1145/3527631} & Extractive & 4,416 articles and 22,057 questions & Health News\\ \hline
        UIT-ViCov19QA \cite{thai2022uit} & Query &  4,500 question-answer pairs & Community FAQ health websites \\ \hline
        VIMQA \cite{le-etal-2022-vimqa} & Bool and Extractive (Multi-hop) & 10,047 questions & Wikipedia (open domain) \\  \hline
        Legal text QA \cite{kien-etal-2020-answering}& Retrieval & 5,922 Vietnamese legal questions & Legal documents \\ \hline
        MLQA \cite{lewis-etal-2020-mlqa} & Extractive & 12,738 questions and answers in English, and 5,029 questions and answers in other languages including Vietnamese & Wikipedia (multilingual open domain) \\ \hline
        HUFI-PostGrad \cite{10.1145/3566123} & Extractive & 45 textual paragraphs and 1,571 question-answer pairs & Closed domain (Regulations on postgraduate admission) \\ \hline
        ViMMRC 2.0 (our) & Multiple choice & 599 passages and 5,273 questions & Students Textbook \\ 
        \bottomrule
    \end{tabular}
    }
\end{table}

Besides, we are willing to introduce different datasets for the machine reading comprehension task for the Vietnamese language. The ViCoQA \cite{10.1007/978-3-030-88113-9_44} dataset is a conversation reading comprehension dataset about health articles from Vietnamese online news. This dataset has 2,000 conversations with 10K question-answer pairs about health. On the other hand, the UIT-ViCoV19QA \cite{thai2022uit} and ViHealthQA \cite{10.1007/978-3-031-10986-7_30} are two datasets constructed based on the community question answering data about the COVID-19 information. The difference between the community-based QA with the extractive MRC is the answers are extracted based on the previously available question-answer pairs. Moreover, we introduce the UIT-EVJVQA \cite{vlsp2022} - a visual QA dataset that combines texts and images to find the correct answers. To have an overview of available Vietnamese datasets for MRC, Table \ref{tab_vi_dataset} shows the information about datasets used for the Vietnamese MRC and QA tasks.  

Apart from the Vietnamese, there are several attempts to construct reading comprehension corpora in other languages to enhance the ability of language understanding of computers. For example, the MC-Test \cite{richardson-etal-2013-mctest}, RACE \cite{lai-etal-2017-race}, CLOTH \cite{xie-etal-2018-large}, OpenBookQA \cite{mihaylov-etal-2018-suit}, ARC \cite{clark2018think}, and DREAM \cite{sun-etal-2019-dream} in English, ASER \cite{10.1145/3579047}, Arabic WikiReading and KaifLematha datasets \cite{albilali2022constructing} in Arabic, the $C^{3}$ dataset \cite{sun-etal-2020-investigating} in Chinese, the Bulgarian multiple-choice reading comprehension dataset \cite{hardalov-etal-2019-beyond} in Bulgarian, the ParsiNLU \cite{khashabi-etal-2021-parsinlu} and PQuAD \cite{DARVISHI2023101486} in Persian, and the BELEBELE \cite{bandarkar2023belebele} - a multilingual multiple-choice reading comprehension dataset with serve for 122 languages. The available dataset is valuable for constructing and evaluating the language comprehension ability of MRC models, especially for low-resource languages.

\subsection{BERTology models}
According to \cite{10.1145/3373017.3373028}, the pre-trained language models have potential advantages since it is trained on a large dataset and can capture the generic features. These features are then utilized for a specific task, such as machine reading comprehension and sentiment analysis. In addition, the pre-trained model can solve the problem of overfitting when training on a limited dataset, especially for low-resource languages. Hence, we chose the pre-trained model to construct the learning model for the multiple-choice reading comprehension task.

BERT \cite{devlin-etal-2019-bert} and its variances play a vital role in many of NLP downstream tasks in recent years. In Vietnamese, the BERTology models also achieve better results than deep learning and traditional models \cite{9352127,to-etal-2021-monolingual,9701541}. The authors in \cite{9352127} applied the multilingual BERT (m-BERT) pre-trained model for multiple-choice reading comprehension in the Vietnamese dataset and received the best results. However, the authors only used the m-BERT, while there are more robust BERTology pre-trained models such as XLM-R \cite{conneau-etal-2020-unsupervised} and DistilBERT \cite{https://doi.org/10.48550/arxiv.1910.01108}. Besides, the authors in \cite{https://doi.org/10.48550/arxiv.2209.10482} take a detailed survey about the performance of BERTology models on various Vietnamese datasets for the text classification task, including the multilingual and monolingual models. It can be seen that BERTology is a robust approach for the multiple-choice reading comprehension task. 

Finally, Jin et al. \cite{jin2020mmm} introduce a multi-stage learning method called the MMM modules that combines the natural language inference (NLI) task with the multiple-choice reading comprehension model to improve performance. The results showed that the performance improved when using the NLI. Moreover, Huynh et al. \cite{huynh-etal-2022-vinli} constructed a large-scale dataset for Vietnamese natural language inference. Therefore, taking the idea from the MMM method \cite{jin2020mmm}, we apply the BERTology models on our new dataset with the NLI as the extra modules to boost the accuracy for the Vietnamese multiple-choice reading comprehension task. 
\section{The Dataset}
\label{dataset}
\subsection{Corpus Creation}
\begin{figure}[H]
    \centering
    \frame{\includegraphics[width=\textwidth]{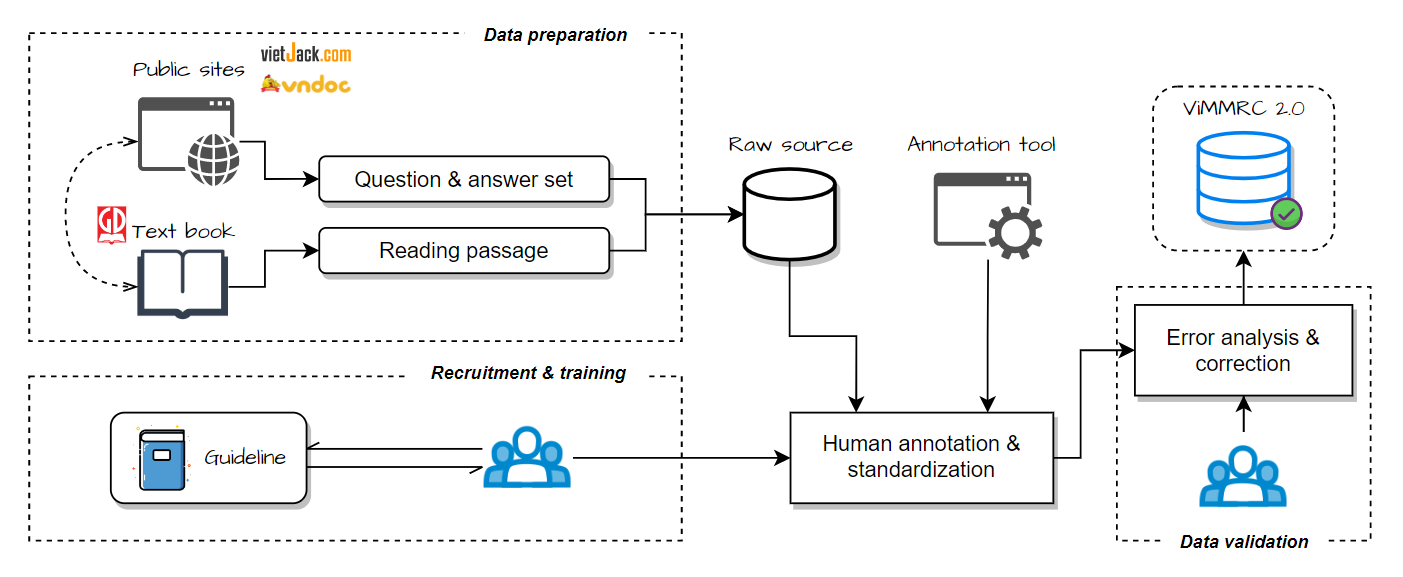}}
    \caption{Corpus creating process for ViMMRC 2.0}
    \label{fig_corpus_create_process}
\end{figure}

\begin{figure}[H]
        \centering
        \frame{\includegraphics[width=\textwidth]{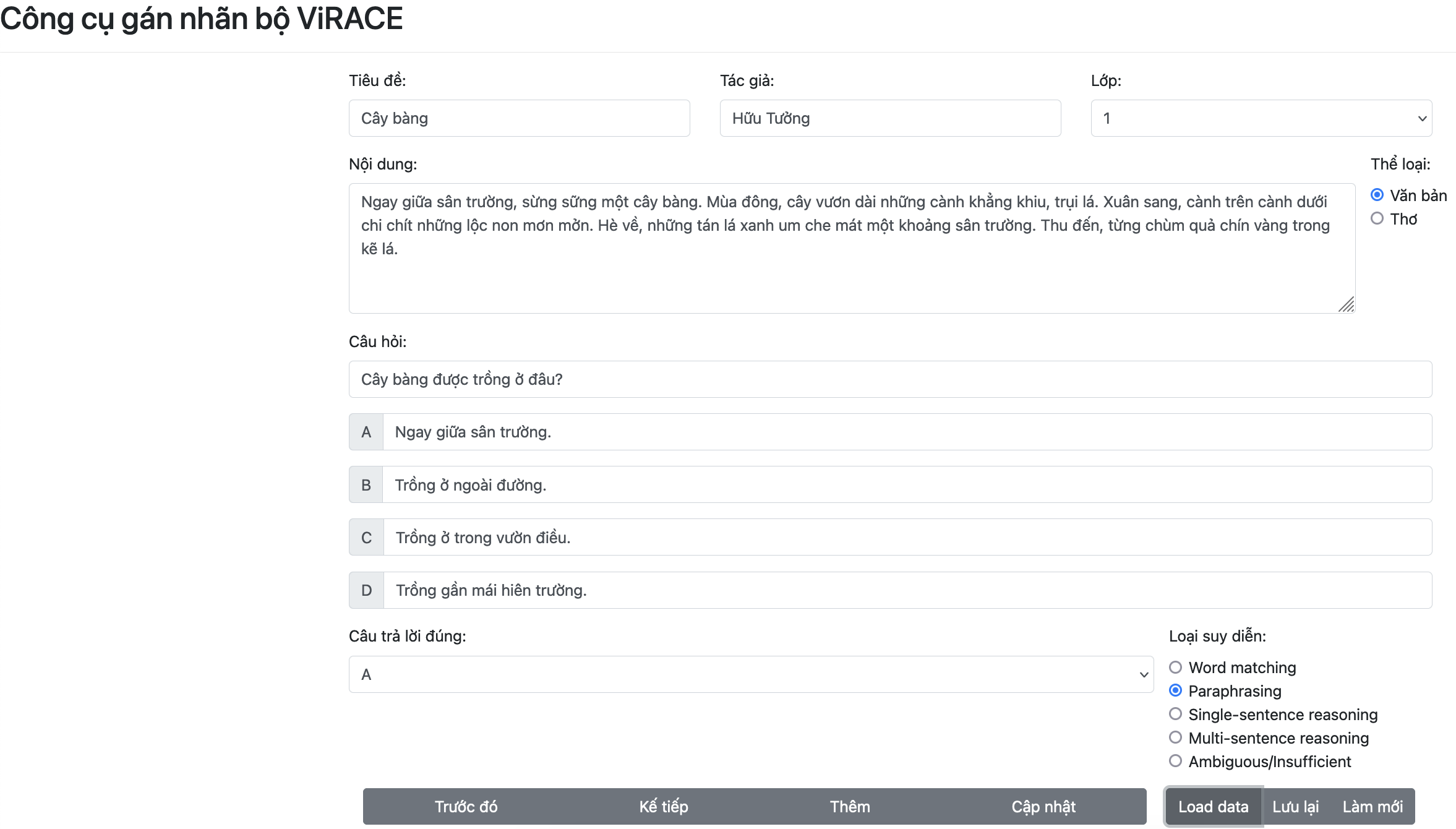}}
        \caption{The Annotation Tool for the ViMMRC multiple-choice reading comprehension dataset}
        \label{fig_annotation_webui}
\end{figure}

Figure \ref{fig_corpus_create_process} illustrates the process of constructing the ViMMRC 2.0 dataset. In this new version, we enhance the number of reading passages and questions based on the previous version. We refer to the data annotation procedure in \cite{finlayson2017overview} to construct the ViMMRC 2.0 corpus. Our steps for creating the ViMMRC 2.0 are described below:

\begin{itemize}
    \item \textbf{Step 1}: Collecting the reading passages and questions from the Vietnamese Literature Student Textbook from $1^{st}$ to $12^{th}$ grades. In addition, we also gather data from two popular free public websites in Vietnam\footnote{vndoc.com and vietjack.com} that provide vast libraries of homework and examinations. 
    \item \textbf{Step 2}: We hire a group of five Vietnamese native speakers to create the dataset for multiple-choice reading comprehension. Fortunately, the collected reading passages have the available solution for each question since the annotators verify the correct choices and import them into the annotation system. All annotators are undergraduate students which all have completed their High School Degrees (The Grade $1^{st}$ to $12^{th}$ in Vietnam).   
    \item \textbf{Step 3}: We build annotation tools with a GUI interface to help annotators create the dataset. Figure \ref{fig_annotation_webui} shows the WebUI for our annotation tools. The tools export the annotated files as JSON files after the annotators finish their work.
    \item \textbf{Step 4}: We let the annotator annotate the data via the annotation tool and send the annotated files via email. After this step, we have a list of annotated JSON files. Each annotator must give the following information: the reading passages, the question, the maximum of four choices for each question, and the correct choice. Besides, for each question in the reading passage, the annotator must give the reasoning types (followed by the previous work in \cite{9247161}). In addition, we ask the annotators to identify whether the reading passage is prose or poem. For each reading passage, annotators must give at least 1 question and answer choice. Every question must have one correct choice.

    \item \textbf{Step 5}: We validate the annotated files and ask the annotators to re-annotate to correct the mistakes in the annotation process. The mistake includes syntax errors in text, no correct choice chosen, and the empty string of questions and answers on each choice. Besides, we also check whether the form of reading passages is prose or poem, and the reasoning types. If the form of passages and the reasoning types are not matched among annotators, we make the final choice by major voting. 
\end{itemize}

Table \ref{tab_sample} shows a sample in the ViMMRC 2.0 dataset. Each reading passage has at least one question, the choices, and the correct choice as answers. In addition, each passage is defined as whether prose or poem by the \textbf{Is prose} attribute. The \textbf{Is prose} is true if it is prose, and is false if it is a poem. 

\begin{table}[H]
    \centering
    \caption{A sample of reading passage in the ViMMRC 2.0 dataset}
    \label{tab_sample}
    \resizebox{\textwidth}{!}{
    \begin{tabular}{lp{15cm}}
        \toprule
        Reading passages & 
        \textbf{Vietnamese}: Trưa mùa hè, nắng vàng như mật ong trải nhẹ trên khắp các cánh đồng cỏ. Những con sơn ca đang nhảy nhót trên sườn đồi. Chúng bay lên cao và cất tiếng hót. Tiếng hót lúc trầm, lúc bổng, lảnh lót vang mãi đi xa. Bỗng dưng lũ sơn ca không hót nữa mà bay vút lên nền trời xanh thẳm.
        (\emph{\textbf{English}: On a summer noon, the golden honey sunlight spreads gently across the grasslands. Larks are dancing on the hillside. They fly up high and start to sing. The sound of low-pitched, high-pitched singing goes on and on. Suddenly, the larks stopped singing and flew up into the deep blue sky.})\\
        \hline
        Is prose & True \\
        \hline
        Question & 
        Những con sơn ca đang nhảy nhót ở đâu?
        \emph{(Where are the larks dancing?)}
        
        A. Trên cánh đồng lúa. 
        \emph{(in the rice field)}
        
        B. Trên sườn đồi. 
        \emph{(on the hillside)}
        
        C. Trên mái hiên nhà. 
        \emph{(on the porch)}
        
        D. Trên đồng cỏ bao la. 
        \emph{(on the vast grasslands)}\\
        Answer & B \\
        \hline
        Question & 
        Thời tiết trong bài được được miêu tả như thế nào? 
        \emph{(How is the weather described in the passage?)}
        
        A. Nắng hanh vàng như chuối. 
        \emph{(Dry and sunny as yellow banana)}

        B. Nắng vàng và trời xanh thẳm. 
        \emph{(Golden sunlight and deep blue sky)}
        
        C. Nóng ôi bức trong người. 
        \emph{(Hot sweltering inside)}
        
        D. Cái nắng lảnh lót. 
        \emph{(Pleasant shining)}\\
        Answer & B\\
        \bottomrule
    \end{tabular}
    }
\end{table}

\subsection{Corpus Overall Statistic}
Table \ref{tab_statis_data} describes the overview information about the dataset. The dataset is divided into the training, development, and test sets with proportion 7-1-2. Compared to the ViMMRC 1.0 \cite{9247161}, the total number of reading passages and questions are improved, which are 699 passages and 5,273 questions for the ViMMRC 2.0 (The previous ViMMRC has 417 reading passages and 2,783 questions). In addition, the vocabulary size also increased. Table \ref{tab_statis_grade} illustrates the number of questions and reading passages by each grade. The vocabulary size is calculated based on the number of tokens in the articles. Unlike the previous dataset, the number of grades is enhanced from $6^{th}$ grade to $12^{th}$ grade. 

\begin{table}[H]
    \centering
    \caption{The overview statistic about the ViMMRC 2.0 dataset}
    \label{tab_statis_data}
    \begin{tabular}{ccccc}
        \toprule
        & \textbf{Train} & \textbf{Dev} & \textbf{Test} & \textbf{All} \\ 
        \midrule 
        Number of passages & 482 & 70 & 147 & 699 \\
        Number of question & 3,600 & 564 & 1,109 & 5,273 \\
        Number of poems & 124 & 13 & 34 & 171 \\
        Number of answers & 14,400 & 2,256 & 4,436 & 21,092 \\
        Vocabulary size & 18,382 & 6,663 & 10,060 & 22,259 \\
        \bottomrule
    \end{tabular}
\end{table}

\begin{figure}[H]
        \centering
        \frame{\includegraphics[width=\textwidth]{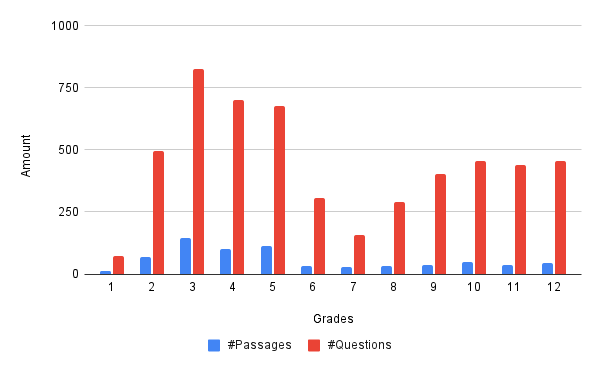}}
        \caption{The distribution of reading passages and questions by each gradge}
        \label{fig_number_question_answer}
\end{figure}

According to Table \ref{tab_statis_grade}, the number of reading passages and questions in Grade $3^{rd}$, Grade $4^{th}$, and Grade $5^{th}$ are more than the rest. In general, according to Figure \ref{fig_number_question_answer}, the number of reading passages and questions in primary grades (Grade $1^{st}$, Grade $5{th}$) is more than the high grades (Grade $6^{st}$, Grade $12{th}$)).

\begin{table}[H]
    \centering
    \caption{Statistical information of the dataset by each grade}
    \label{tab_statis_grade}
    \begin{tabular}{cccccc}
        \toprule
        \textbf{Grades} & \textbf{\#Passages} & \textbf{\#Questions} & \textbf{\#Answers} & \textbf{\#Poems} & \textbf{Vocab. size} \\
        \midrule
        $1^{st}$ & 11 & 71 & 284 & 1 & 563 \\
        $2^{nd}$ & 69 & 497 & 1,988 & 11 & 3,192 \\
        $3^{rd}$ & 144 & 826 & 3,304 & 32 & 5,110 \\
        $4^{th}$ & 102 & 699 & 2,796 & 16 & 4,901 \\
        $5^{th}$ & 111 & 675 & 2,700 & 18 & 5,798 \\
        $6^{th}$ & 33 & 308 & 1,232 & 3 & 6,155 \\
        $7^{th}$ & 29 & 158 & 632 & 14 & 4,439 \\
        $8^{th}$ & 32 & 292 & 1,168 & 7 & 6,550 \\
        $9^{th}$ & 38 & 401 & 1,604 & 18 & 6,390 \\
        $10^{th}$ & 47 & 454 & 1,816 & 19 & 6,778 \\
        $11^{th}$ & 37 & 438 & 1,752 & 21 & 7,253 \\
        $12^{th}$ & 46 & 454 & 1,816 & 11 & 8,646 \\
        \bottomrule
    \end{tabular}
\end{table}

\begin{table}[H]
    \centering
    \caption{Overall statistic of the ViMMRC 2.0 dataset on the training, development, and test sets}
    \label{tab_overall_stat}
    \resizebox{\textwidth}{!}{
    \begin{tabular}{lrrrrrrrrrrrr}
    \toprule
    \multicolumn{13}{c}{\textbf{Training set}}                 \\ \hline
    \multicolumn{1}{l}{\textit{Grade}}                     & \multicolumn{1}{r}{\textit{\textbf{$1^{st}$}}} & \multicolumn{1}{r}{\textit{\textbf{$2^{nd}$}}} & \multicolumn{1}{r}{\textit{\textbf{$3^{rd}$}}} & \multicolumn{1}{r}{\textit{\textbf{$4^{th}$}}} & \multicolumn{1}{r}{\textit{\textbf{$5^{th}$}}} & \multicolumn{1}{r}{\textit{\textbf{$6^{th}$}}} & \multicolumn{1}{r}{\textit{\textbf{$7^{th}$}}} & \multicolumn{1}{r}{\textit{\textbf{$8^{th}$}}} & \multicolumn{1}{r}{\textit{\textbf{$9^{th}$}}} & \multicolumn{1}{r}{\textit{\textbf{$10^{th}$}}} & \multicolumn{1}{r}{\textit{\textbf{$11^{th}$}}} & \multicolumn{1}{r}{\textit{\textbf{$12^{th}$}}} \\ 
    \multicolumn{1}{l}{\textit{\# Poem}}     & \multicolumn{1}{r}{1}                   & \multicolumn{1}{r}{9}                   & \multicolumn{1}{r}{20}                  & \multicolumn{1}{r}{12}                  & \multicolumn{1}{r}{11}                  & \multicolumn{1}{r}{3}                   & \multicolumn{1}{r}{13}                  & \multicolumn{1}{r}{6}                   & \multicolumn{1}{r}{17}                  & \multicolumn{1}{r}{14}                   & \multicolumn{1}{r}{12}                   & 6                                         \\ 
    \multicolumn{1}{l}{\textit{\# Passage}}  & \multicolumn{1}{r}{7}                   & \multicolumn{1}{r}{48}                  & \multicolumn{1}{r}{99}                  & \multicolumn{1}{r}{71}                  & \multicolumn{1}{r}{77}                  & \multicolumn{1}{r}{23}                  & \multicolumn{1}{r}{20}                  & \multicolumn{1}{r}{22}                  & \multicolumn{1}{r}{26}                  & \multicolumn{1}{r}{32}                   & \multicolumn{1}{r}{25}                   & 32                                        \\ 
    \multicolumn{1}{l}{\textit{\# Question}} & \multicolumn{1}{r}{44}                  & \multicolumn{1}{r}{341}                 & \multicolumn{1}{r}{563}                 & \multicolumn{1}{r}{501}                 & \multicolumn{1}{r}{456}                 & \multicolumn{1}{r}{219}                 & \multicolumn{1}{r}{94}                  & \multicolumn{1}{r}{209}                 & \multicolumn{1}{r}{271}                 & \multicolumn{1}{r}{331}                  & \multicolumn{1}{r}{273}                  & 318                                       \\ 
    \multicolumn{1}{l}{\textit{\# Answers}}  & \multicolumn{1}{r}{176}                 & \multicolumn{1}{r}{1,364}                & \multicolumn{1}{r}{2,252}                & \multicolumn{1}{r}{2,004}                & \multicolumn{1}{r}{1,824}                & \multicolumn{1}{r}{876}                 & \multicolumn{1}{r}{376}                 & \multicolumn{1}{r}{836}                 & \multicolumn{1}{r}{1,084}                & \multicolumn{1}{r}{1,244}                 & \multicolumn{1}{r}{1,092}                 & 1,272                                      \\ 
    \multicolumn{1}{l}{\textit{Vocab size}}  & \multicolumn{1}{r}{319}                 & \multicolumn{1}{r}{2,470}                & \multicolumn{1}{r}{4,165}                & \multicolumn{1}{r}{3,994}                & \multicolumn{1}{r}{4,814}                & \multicolumn{1}{r}{5,054}                & \multicolumn{1}{r}{3,165}                & \multicolumn{1}{r}{5,203}                & \multicolumn{1}{r}{4,253}                & \multicolumn{1}{r}{5,834}                 & \multicolumn{1}{r}{5,827}                 & 7,304                                      \\ \hline
    \multicolumn{13}{c}{\textbf{Development set}}                                                                               \\ \hline
    \multicolumn{1}{l}{\textit{Grade}}                     & \multicolumn{1}{r}{\textit{\textbf{$1^{st}$}}} & \multicolumn{1}{r}{\textit{\textbf{$2^{nd}$}}} & \multicolumn{1}{r}{\textit{\textbf{$3^{rd}$}}} & \multicolumn{1}{r}{\textit{\textbf{$4^{th}$}}} & \multicolumn{1}{r}{\textit{\textbf{$5^{th}$}}} & \multicolumn{1}{r}{\textit{\textbf{$6^{th}$}}} & \multicolumn{1}{r}{\textit{\textbf{$7^{th}$}}} & \multicolumn{1}{r}{\textit{\textbf{$8^{th}$}}} & \multicolumn{1}{r}{\textit{\textbf{$9^{th}$}}} & \multicolumn{1}{r}{\textit{\textbf{$10^{th}$}}} & \multicolumn{1}{r}{\textit{\textbf{$11^{th}$}}} & \multicolumn{1}{r}{\textit{\textbf{$12^{th}$}}} \\ 
    \multicolumn{1}{l}{\textit{\# Poem}}     & \multicolumn{1}{r}{0}                   & \multicolumn{1}{r}{0}                   & \multicolumn{1}{r}{4}                   & \multicolumn{1}{r}{0}                   & \multicolumn{1}{r}{2}                   & \multicolumn{1}{r}{0}                   & \multicolumn{1}{r}{0}                   & \multicolumn{1}{r}{0}                   & \multicolumn{1}{r}{0}                   & \multicolumn{1}{r}{1}                    & \multicolumn{1}{r}{4}                    & 2                                         \\ 
    \multicolumn{1}{l}{\textit{\# Passage}}  & \multicolumn{1}{r}{1}                   & \multicolumn{1}{r}{7}                   & \multicolumn{1}{r}{15}                  & \multicolumn{1}{r}{10}                  & \multicolumn{1}{r}{11}                  & \multicolumn{1}{r}{3}                   & \multicolumn{1}{r}{3}                   & \multicolumn{1}{r}{3}                   & \multicolumn{1}{r}{4}                   & \multicolumn{1}{r}{5}                    & \multicolumn{1}{r}{4}                    & 4                                         \\ 
    \multicolumn{1}{l}{\textit{\# Question}} & \multicolumn{1}{r}{5}                   & \multicolumn{1}{r}{54}                  & \multicolumn{1}{r}{92}                  & \multicolumn{1}{r}{53}                  & \multicolumn{1}{r}{83}                  & \multicolumn{1}{r}{25}                  & \multicolumn{1}{r}{22}                  & \multicolumn{1}{r}{22}                  & \multicolumn{1}{r}{52}                  & \multicolumn{1}{r}{47}                   & \multicolumn{1}{r}{65}                   & 44                                        \\ 
    \multicolumn{1}{l}{\textit{\# Answers}}  & \multicolumn{1}{r}{20}                  & \multicolumn{1}{r}{216}                 & \multicolumn{1}{r}{368}                 & \multicolumn{1}{r}{212}                 & \multicolumn{1}{r}{332}                 & \multicolumn{1}{r}{100}                 & \multicolumn{1}{r}{88}                  & \multicolumn{1}{r}{88}                  & \multicolumn{1}{r}{208}                 & \multicolumn{1}{r}{188}                  & \multicolumn{1}{r}{260}                  & 176                                       \\ 
    \multicolumn{1}{l}{\textit{Vocab size}}  & \multicolumn{1}{r}{74}                  & \multicolumn{1}{r}{727}                 & \multicolumn{1}{r}{1,259}                & \multicolumn{1}{r}{1,246}                & \multicolumn{1}{r}{1,367}                & \multicolumn{1}{r}{1,342}                & \multicolumn{1}{r}{1,046}                & \multicolumn{1}{r}{1,063}                & \multicolumn{1}{r}{2,959}                & \multicolumn{1}{r}{1,244}                 & \multicolumn{1}{r}{754}                  & 1,555                                      \\ \hline
    \multicolumn{13}{c}{\textbf{Test set}}                                                                                               \\ \hline
    \multicolumn{1}{l}{\textit{Grade}}                     & \multicolumn{1}{r}{\textit{\textbf{$1^{st}$}}} & \multicolumn{1}{r}{\textit{\textbf{$2^{nd}$}}} & \multicolumn{1}{r}{\textit{\textbf{$3^{rd}$}}} & \multicolumn{1}{r}{\textit{\textbf{$4^{th}$}}} & \multicolumn{1}{r}{\textit{\textbf{$5^{th}$}}} & \multicolumn{1}{r}{\textit{\textbf{$6^{th}$}}} & \multicolumn{1}{r}{\textit{\textbf{$7^{th}$}}} & \multicolumn{1}{r}{\textit{\textbf{$8^{th}$}}} & \multicolumn{1}{r}{\textit{\textbf{$9^{th}$}}} & \multicolumn{1}{r}{\textit{\textbf{$10^{th}$}}} & \multicolumn{1}{r}{\textit{\textbf{$11^{th}$}}} & \multicolumn{1}{r}{\textit{\textbf{$12^{th}$}}} \\
    \multicolumn{1}{l}{\textit{\# Poem}}     & \multicolumn{1}{r}{0}                   & \multicolumn{1}{r}{2}                   & \multicolumn{1}{r}{8}                   & \multicolumn{1}{r}{4}                   & \multicolumn{1}{r}{5}                   & \multicolumn{1}{r}{0}                   & \multicolumn{1}{r}{1}                   & \multicolumn{1}{r}{1}                   & \multicolumn{1}{r}{1}                   & \multicolumn{1}{r}{4}                    & \multicolumn{1}{r}{5}                    & 3                                         \\ 
    \multicolumn{1}{l}{\textit{\# Passage}}  & \multicolumn{1}{r}{3}                   & \multicolumn{1}{r}{14}                  & \multicolumn{1}{r}{30}                  & \multicolumn{1}{r}{21}                  & \multicolumn{1}{r}{23}                  & \multicolumn{1}{r}{7}                   & \multicolumn{1}{r}{6}                   & \multicolumn{1}{r}{7}                   & \multicolumn{1}{r}{8}                   & \multicolumn{1}{r}{10}                   & \multicolumn{1}{r}{8}                    & 10                                        \\ 
    \multicolumn{1}{l}{\textit{\# Question}} & \multicolumn{1}{r}{22}                  & \multicolumn{1}{r}{102}                 & \multicolumn{1}{r}{171}                 & \multicolumn{1}{r}{145}                 & \multicolumn{1}{r}{136}                 & \multicolumn{1}{r}{64}                  & \multicolumn{1}{r}{42}                  & \multicolumn{1}{r}{61}                  & \multicolumn{1}{r}{78}                  & \multicolumn{1}{r}{96}                   & \multicolumn{1}{r}{100}                  & 92                                        \\ 
    \multicolumn{1}{l}{\textit{\# Answers}}  & \multicolumn{1}{r}{88}                  & \multicolumn{1}{r}{408}                 & \multicolumn{1}{r}{684}                 & \multicolumn{1}{r}{580}                 & \multicolumn{1}{r}{544}                 & \multicolumn{1}{r}{256}                 & \multicolumn{1}{r}{168}                 & \multicolumn{1}{r}{244}                 & \multicolumn{1}{r}{312}                 & \multicolumn{1}{r}{384}                  & \multicolumn{1}{r}{400}                  & 368                                       \\ 
    \multicolumn{1}{l}{\textit{Vocab size}}  & \multicolumn{1}{r}{243}                 & \multicolumn{1}{r}{1,162}                & \multicolumn{1}{r}{2,054}                & \multicolumn{1}{r}{1,879}                & \multicolumn{1}{r}{2,080}                & \multicolumn{1}{r}{2,058}                & \multicolumn{1}{r}{1,999}                & \multicolumn{1}{r}{2,509}                & \multicolumn{1}{r}{2,818}                & \multicolumn{1}{r}{2,233}                 & \multicolumn{1}{r}{3,086}                 & 3,668                                      \\
    \bottomrule
    \end{tabular}
    }
\end{table}

Finally, Table \ref{tab_overall_stat} shows an overview of the number of reading passages, questions, and answers for each grade in the training, development, and test sets. It can be seen that the distribution of the training, development, and test sets are similar. 

    

\subsection{Reading Passages Analysis}
\begin{figure}[H]
        \centering
        \frame{\includegraphics[width=.9\textwidth]{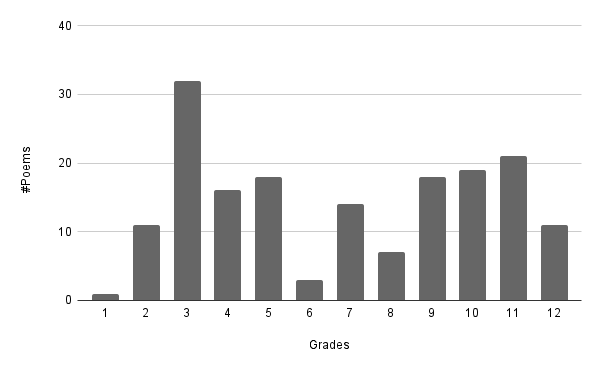}}
        \caption{The number of poems as reading passages by each gradge}
        \label{fig_number_poems}
\end{figure}

According to Figure \ref{fig_number_poems}, the number of poems in higher grades is equally distributed between the primary and high grades. Grades $3^{rd}$ has the largest number of poems as reading passages, while Grades $1^{rd}$ have only one poem, according to Table \ref{tab_statis_grade}. 

\begin{table}[H]
    \centering
    \caption{Comparison between the prose and poem in the ViMMRC 2.0 dataset}
    \label{tab_prose_poem}
    \begin{tabular}{lcc}
        \toprule
        & \textbf{Proses} & \textbf{Poems} \\
        \midrule
        Number of passages & 528 & 171 \\
        Average length of reading passages & 2,826.57 & 1,035.03 \\
        Vocabulary size & 20,506 & 7,318 \\
        \bottomrule
    \end{tabular}
\end{table}

On the other hand, Table \ref{tab_prose_poem} shows the statistical comparison between the reading passages in prose form and poem form, respectively. The average length of articles is calculated by the sum of the length of articles divided by the number of articles. According to Table \ref{tab_prose_poem}, the average length of articles in the poem form is shorter than in the prose form. Also, the vocabulary size in the poem also less than in the prose. It is clear that the articles in prose are longer and have more words than those in poems.  

\begin{figure}[H]
        \centering
        \frame{\includegraphics[width=.9\textwidth]{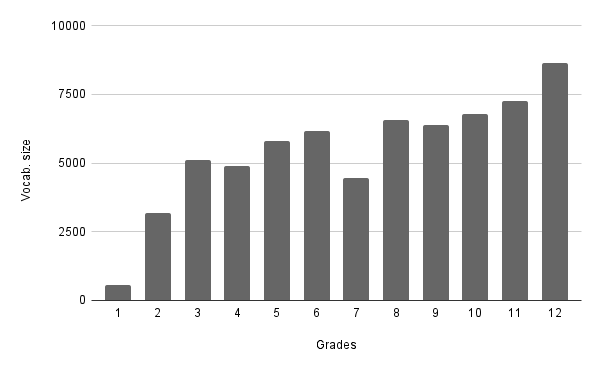}}
        \caption{The vocabulary size by each grade}
        \label{fig_vocab}
\end{figure}

Finally, according to Figure \ref{fig_vocab}, the vocabulary size is increased from the low grades to the high grades. This indicates that the difficulty of reading passages is increased grade by grade. 

\subsection{Questions Analysis}
In this section, we analyze the types of questions in the ViMMRC 2.0 dataset. There are three types of questions: Wh-questions, listed questions, and Yes/No questions, according to \cite{phan2010question}. Besides, several questions are not explicitly classified into those three types of questions, so we categorize them as Other. The types and samples of questions are described in Table \ref{tab_ques_type}. The correct options are marked as bold text. We manually annotated these types of questions by humans on the development set. 

\begin{table}[ht!]
    \centering
    \caption{Types of questions in the ViMMRC 2.0 datasets}
    \label{tab_ques_type}
    \resizebox{\textwidth}{!}{
    \begin{tabular}{cp{5cm}p{15cm}}
        \toprule
        \textbf{Type} & \textbf{Description} & \textbf{Samples} \\
        \midrule
        1 & \makecell[{{p{5cm}}}]{Wh-Questions. This type of questions is about What (cái gì), Who (ai, người nào), Which (vật gì, con gì), How (như thế nào), and How many (Bao nhiêu). The correct option is often matched directly from the reading passages.} & \makecell[{{p{15cm}}}]{Question: Bố Dũng đến trường học để làm điều gì? \emph{(English: What does Dung's Father do when he goes to school with his son?)} \\ 
        A. Để họp phụ huynh đầu năm \emph{(To join the parental meeting)}. \\
        B. Để đưa đồ cho Dũng \emph{(To give things to Dung)}. \\
        \textbf{C. Để chào người thầy giáo cũ \emph{(To visit an old teacher)}.} \\
        D. Để đưa Dũng đi học \emph{(To take Dung to his school)}. \\
        } \\
        \hline
        2 & \makecell[{{p{5cm}}}]{Questions with listed options. This type of question gives a list of options and asks people to choose the correct one, all of the above, or none of the above. To choose the correct options, people need to read a bunch of paragraphs or even all the passages.} & \makecell[{{p{15cm}}}]{
        Question: Ý nghĩa của truyện Sơn Tinh, Thủy Tinh là gì? \emph{(English: What is the meaning of Son Tinh Thuy Tinh folktale?)} \\
        A. Giải thích hiện tượng lũ lụt ở nước ta hằng năm \emph{(To explain the appearance of storm and flood)} \\
        B. Thể hiện ước nguyện của con người trong việc chế ngự thiên nhiên. \emph{(To show the dream of controlling the disaster of humans)}. \\
        C. Ca ngợi công lao dựng nước của các vua Hùng. \emph{(To honor the feat of Hùng kings in the construction of the nation)} \\ 
        \textbf{D. Cả A, B và C đều đúng. \emph{(All of the choices are correct)}
        }\\}\\
        \hline
        3 & \makecell[{{p{5cm}}}]{Yes/No questions. This question have only two options: Yes (Có / Đúng) or No (Không / Sai).} &  \makecell[{{p{15cm}}}]{
        Question: Cốt truyện của Lặng lẽ Sa Pa là cốt truyện có tính kịch tính, xung đột. Đúng hay sai? \emph{(English: The story of Lang le Sa pa is a dramatic story, isn't it?)} \\
        A. Đúng \emph{(Yes)}. \\
        \textbf{B. Sai \emph{(No)}.} \\
        }\\
        \hline
        4 & \makecell[{{p{5cm}}}]{Others} & \makecell[{{p{15cm}}}]{Question 1: Hãy sắp xếp các chi tiết dưới đây theo đúng thứ tự xuất hiện trong truyện Sơn Tinh, Thủy Tinh.
        \emph{(English: Let arrange these facts from the Son Tinh Thuy Tinh story by order of appearance})\\
        1. Hùng Vương thứ mười tám nêu ra yêu cầu về lễ vật. \emph{(Hùng King XVIII gives the requirements of the marriage gift)} \\
        2. Sơn Tinh đem lễ vật đến trước và cưới được vợ. \emph{(Son Tinh comes first and married the bride)} \\
        3. Vua Hùng tổ chức kén rể cho Mị Nương. \emph{(Hùng King XVIII organizes a wedding for his daughter)} \\
        4. Sơn Tinh – Thủy Tinh đánh nhau ròng rã mấy tháng trời. \emph{(Son Tinh and Thuy Tinh keep fighting for months)} \\
        A. (1) - (2) - (3) - (4). \\
        B. (1) - (3) - (2) - (4). \\
        \textbf{C. (3) - (1) - (2) - (4).} \\
        D. (1) - (3) - (4) - (2). \\
        Question 2: Điền từ chỉ màu sắc thích hợp vào khổ thơ sau:  Em yêu màu ...  Đồng bằng, rừng núi  Biển đầy cá tôm,  Bầu trời cao vợi \emph{(Choose the word that describe the correct color to fill the blank: "Em yêu màu ...  Đồng bằng, rừng núi  Biển đầy cá tôm,  Bầu trời cao vợi")}. \\
        \textbf{A. Xanh \emph{(Blue)}.} \\
        B. Cam \emph{(Orange)}. \\
        C. Trắng \emph{(White)}. \\
        D. Đỏ \emph{(Red)}. \\
        }\\
        \bottomrule
    \end{tabular}
    }
\end{table}


\begin{table}[H]
    \centering
    \caption{Statistic of questions by types in the development set}
    \label{tab_stat_questions}
        \begin{tabular}{clc}
            \toprule
            \textbf{Types} & \textbf{Description} & \textbf{Numbers} \\
            \midrule
            1 & Wh-Questions & 445 \\
            2 & Questions with listed options & 96 \\
            3 & Yes/No questions & 5 \\
            4 & Others & 18 \\
            \bottomrule
        \end{tabular}
\end{table}

\begin{table}[H]
    \centering
    \caption{The number of options in each question in the ViMMRC 2.0 corpus}
        \begin{tabular}{lp{4cm}p{4cm}}
            \toprule
            & \textbf{Has 2 options} & \textbf{Has 3 options} \\
            \midrule
            Training set & 30 & 88 \\
            Development set & 5 & 9 \\
            Test set & 42 & 117 \\
            \bottomrule
        \end{tabular}
        \label{tab_num_options}
\end{table}

\begin{figure}[ht!]
\centering
    \begin{minipage}{0.49\textwidth}
    \centering
    \includegraphics[width=\textwidth]{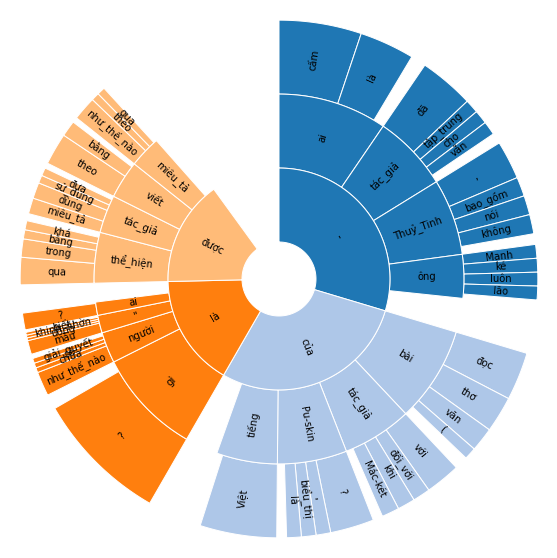}
    \\ 
    (a) ViMMRC 2.0
    \end{minipage}
    \begin{minipage}{0.49\textwidth}
    \centering
    \includegraphics[width=\textwidth]{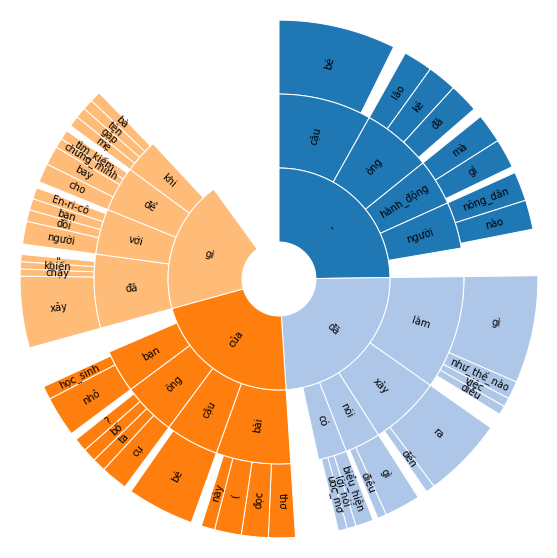}
    \\
    (b) ViMMRC 1.0
    \end{minipage}
    \begin{minipage}{0.49\textwidth}
    \centering
    \includegraphics[width=\textwidth]{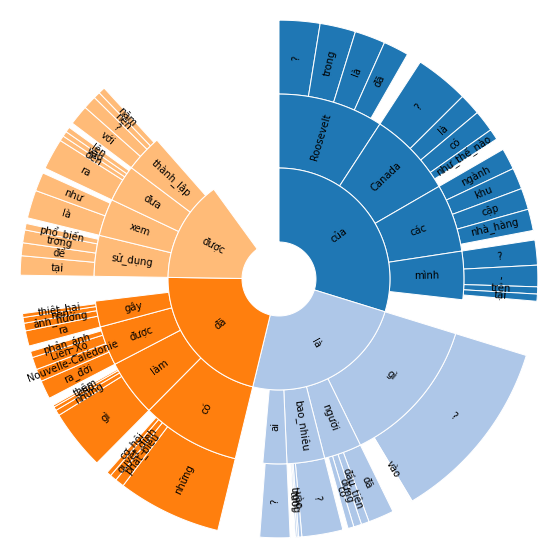}
    \\
    (c) ViQuAD
    \end{minipage}
    \caption{Distribution of trigram prefixes in the ViMMRC 2.0, ViMMRC 1.0, and ViQuAD corpora}
    \label{fig_trigram_distribution_vimmrc}
\end{figure}

In addition, we statistic the number of questions by each of those types in the development set. Table \ref{tab_stat_questions} shows the results of questions by types on the dev set. From Table \ref{tab_stat_questions}, it can be seen that most of the questions are Wh-questions. The number of questions with listed options also accounts large proportion. In contrast, the number of Yes/No questions is very few. Besides, as shown in \ref{tab_ques_type}, the number of options in each question is not always full of four. In some questions, such as the Yes/No questions or listed options questions, the options may be less than four. Therefore, this is different from the previous version of the corpus. Table \ref{tab_num_options} shows the number of options in the training, development, and test sets. The "Other" questions have different characteristics from the remaining types. However, when analyzing several samples from the development set, we found two forms: the questions require the correct answer as a list of mentioned facts from the reading passage in the right order (Question 1 of the Other type in Table \ref{tab_ques_type}), and the questions contain the cloze (Question 2 of the Other type in Table \ref{tab_ques_type}). To give the correct answers, readers must understand the whole passage and link the mentioned facts.

To have an in-depth understanding of the characteristics of questions, we use the n-gram model to explore the distribution of words in the questions. Before conducting the analysis, the questions are tokenized into words by using the VnCoreNLP toolkit \cite{vu-etal-2018-vncorenlp}. According to previous work in \cite{reddy-etal-2019-coqa}, we use the trigram model to construct the distribution graph for questions in the dataset. Figure \ref{fig_trigram_distribution_vimmrc} illustrates the distribution of words in the question on the ViMMRC 2.0, ViMMRC 1.0, and UIT-ViQuAD corpora.

\begin{figure}[ht!]
\centering
    \begin{minipage}{0.49\textwidth}
    \centering
    \includegraphics[width=\textwidth]{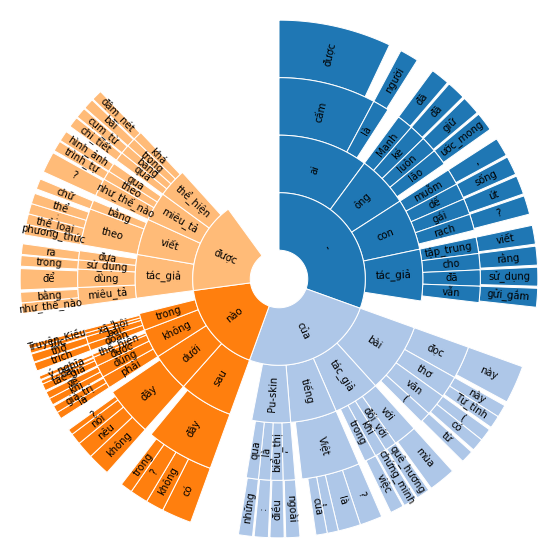}
    \\ 
    (a) WH-questions
    \end{minipage}
    \begin{minipage}{0.49\textwidth}
    \centering
    \includegraphics[width=\textwidth]{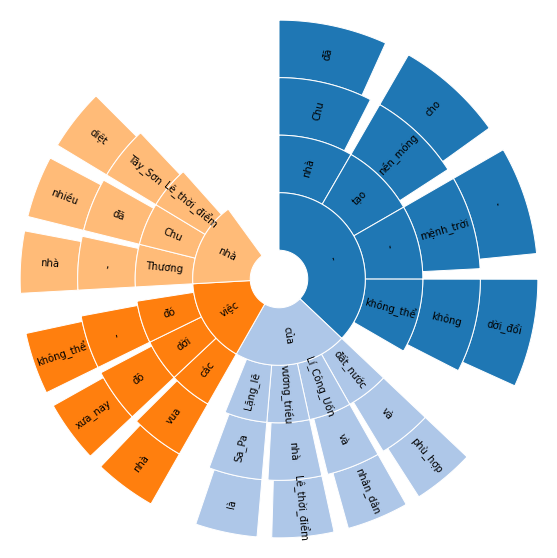}
    \\
    (b) Yes/No questions
    \end{minipage}
    \begin{minipage}{0.49\textwidth}
    \centering
9    \includegraphics[width=\textwidth]{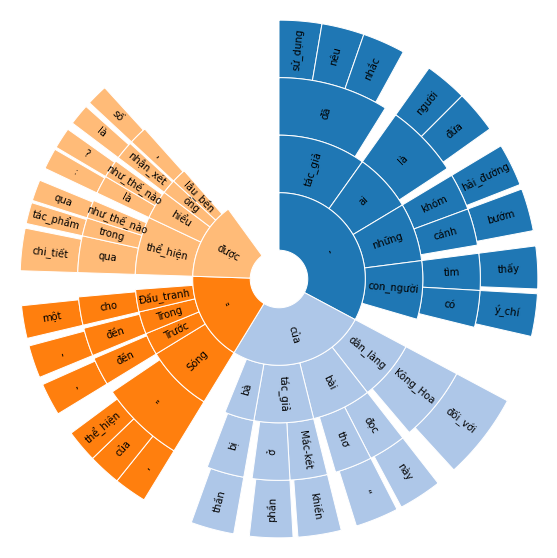}
    \\
    (c) Listed questions
    \end{minipage}
    \begin{minipage}{0.49\textwidth}
    \centering
    \includegraphics[width=\textwidth]{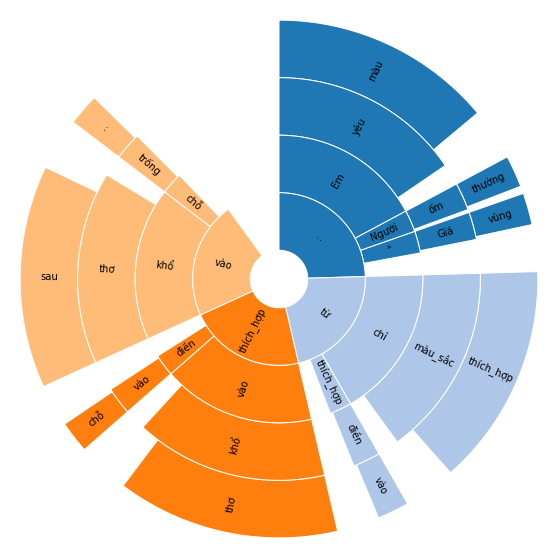}
    \\
    (d) Other questions
    \end{minipage}
    \caption{Distribution of trigram prefixes in each question type in the ViMMRC 2.0 dataset}
    \label{fig_trigram_question_types}
\end{figure}

According to Figure \ref{fig_trigram_distribution_vimmrc}, the questions which contain the word "được" (indicated for How, which), "là" (represented for Who, What), and "của" (represented for Whose, Which) take the same proportion. In the literature textbook, those three words usually appear. For example, the question "Đảo Khỉ là khu vực \textbf{được} miêu tả như thế nào?" (How is Dao Khi described?) or Chứng cứ nào không \textbf{được} tác giả dùng để chứng minh “cái hay” của tiếng Việt? (Which evidence is not used by the author to prove the unique point in the Vietnamese?) contains the word "được" to indicate a question. Especially, questions that have the "được" in Vietnamese are often in the passive form. On the other hand, the word "là" often indicates the Who or Which questions, for example, "Chuyện Lặng lẽ Sa Pa có nhân vật chính \textbf{là} ai?" (Who is the main character in Lang Le Sa Pa story?) and "Yếu tố bất ngờ trong truyện là gì?" (Which is the surprising point in the story?). In contrast with the word "được", the questions with "là" are often in the active form in Vietnamese. Finally, the questions that contain "của" are often in the type of possessive form question, for example, "Tên hiệu \textbf{của} nhà thơ Nguyễn Khuyến là:" (whose other name is poet Nguyen Khuyen?).   

Moreover, to show the characteristic of words in each type of question as shown in Table \ref{tab_ques_type}, Figure \ref{fig_trigram_question_types} illustrates the distribution of the trigram in the development set of the ViMMRC 2.0 by each type. It can be seen that, in the WH-questions, the appearance of the three prefixes "được", "của", and "nào" are the point to identify this type of question. Besides, for the listed questions type, the "được" and "của" frequently appear in the questions. Specifically, the difference between those types is that the question in the WH-question often aims at an object, while the question in the listed option usually generalizes the target. For the Yes/No question, the objects are often mentioned directly in the question, so the phrase "Đúng hay sai" often appears. Finally, for the Other type of questions, the distribution of trigram is not similar to the three remaining types.

\section{Methodology}
\label{method}
\subsection{Baseline Methods}
The multiple-choice reading comprehension task can be denoted as giving a passage P with a question Q. Then the reading comprehension model must give the correct answer in a list of candidate answers.  According to work in \cite{9352127}, the reading passage (P), the question (Q), and the answers (A) are represented as embedding vectors with a specific dimension and length of text. Also, according to \cite{9352127}, the empirical results show that the BERT model for encoding the text has more outstanding results than the Co-matching model. 
Therefore, in this new version of the dataset, we choose the BERT model as our baseline approach to construct the multiple-choice reading comprehension model. 

\begin{table}[ht!]
    \centering
    \caption{Pre-trained BERTology model used on the ViMMRC 2.0 corpus}
    \label{tab_pre_trained}
    \begin{tabular}{ll}
        \toprule
        \textbf{BERT architecture} & \textbf{Pre-trained model} \\
        \midrule
        \multicolumn{2}{c}{\textbf{Multilingual models}} \\
        BERT\cite{devlin-etal-2019-bert} & bert-based-multilingual-cased \\   
        XLM-R\cite{conneau-etal-2020-unsupervised} & xlm-r-based \\
        \hline
        \multicolumn{2}{c}{\textbf{Monolingual models}} \\  
        viBERT\cite{bui-etal-2020-improving} & vibert-base-cased \\    
        BERT4News\cite{https://doi.org/10.48550/arxiv.2101.12672} & vibert4news-base-cased \\
        \bottomrule
    \end{tabular}
\end{table}

In addition, according to \cite{rogers-etal-2020-primer}, BERT and its variances, which were trained on large-scale corpus on various languages, is a standard baseline for many NLP tasks. Referred to \cite{https://doi.org/10.48550/arxiv.2209.10482}, we choose the BERT, XLM-R, viBERT, and BERT4News with relevant pre-trained models for Vietnamese. Table \ref{tab_pre_trained} describes those models and pre-trained in detail. 

\subsection{Multi-stage Learning}
Tan et al. \cite{10.1145/3453185} introduce a novel two-stage framework for BERT that performed well on the multiple-choice task. The two-stage framework consists of two main phases: pre-training and fine-tuning. The pre-training stage constructs a masked language model on a large-scale corpus that covers the domain for other downstream tasks. The fine-tuning task adapts the pre-trained language model to a specific task such as the multiple-choice task. The MMM \cite{jin2020mmm} has a similar idea to the two-stage framework since it employs the Coarse-tuning and Multi-task Learning stages. We follow the MMM framework \cite{jin2020mmm} for multi-stage and multi-task learning for the multiple-choice reading comprehension on the ViMMRC 2.0 dataset. Due to the lack of addition in-domain dataset, we did not use the proportional multi-task learning strategy, but sequential fine-tuning of our model on only the target dataset via the MRC task fine-tuning stage instead. Figure \ref{fig_multi_stage} illustrates the two-stage procedure to train our model. This procedure consists of two main stages as follows:

\begin{figure}[ht]
        \centering
        \includegraphics[width=\textwidth]{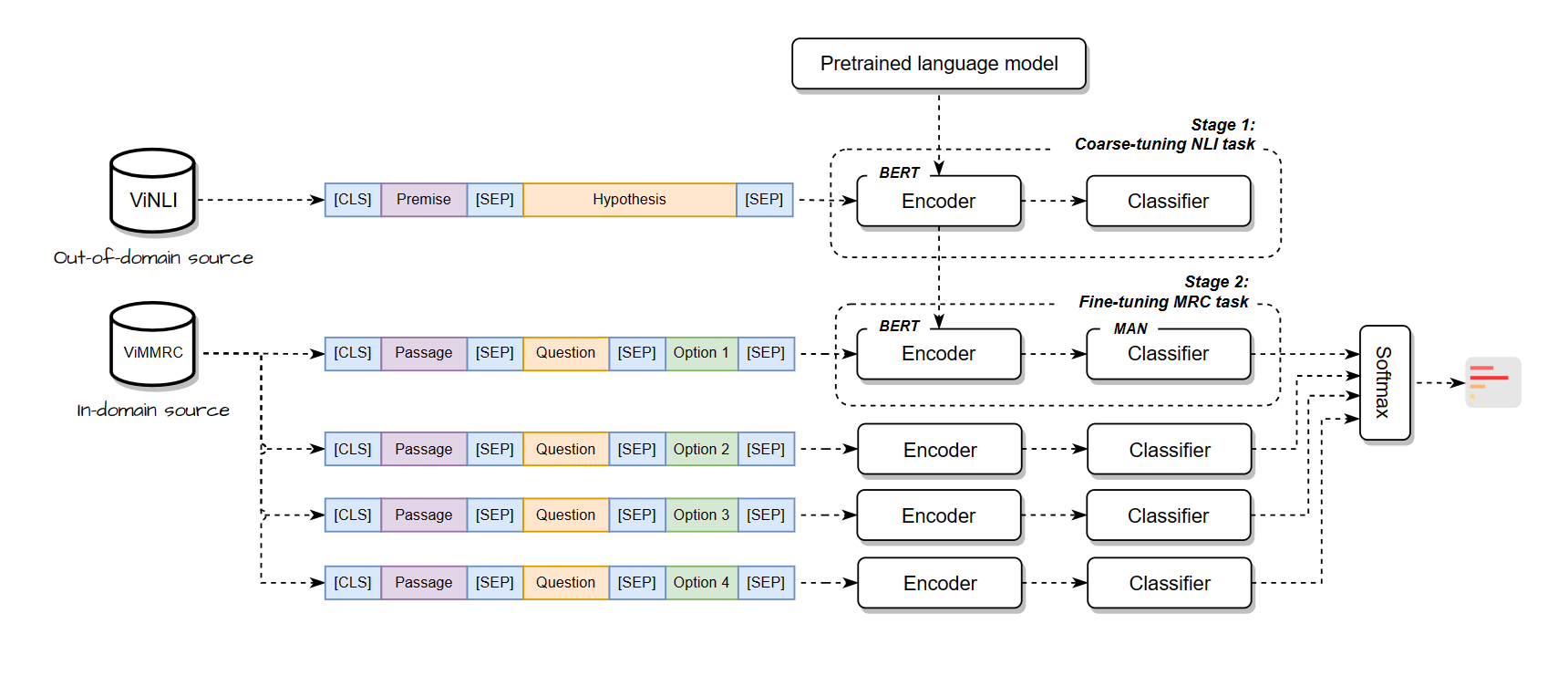}
        \caption{Multi-stage fine-tuning strategy using both in-domain and out-of-domain datasets}
        \label{fig_multi_stage}
\end{figure}

\begin{itemize}
    \item \textbf{Coarse-tuning stage with NLI task}: According to \cite{jin2020mmm}, the NLI task, which is trained on an out-of-domain dataset, shows robustness results in improving the performance of the multiple-choice reading comprehension task. Therefore, we use the ViNLI \cite{huynh-etal-2022-vinli} - an open-domain natural language inference on Vietnamese texts constructed from 800 online news articles with 13 different topics to train the NLI models for coarse tuning in this stage. The ViNLI consists of three labels: \textit{contradiction}, \textit{entailment}, and \textit{neutral} for determining the entailment between the premise and the hypothesis sentences \cite{huynh-etal-2022-vinli}. 
    \item \textbf{Fine-tuning stage with MRC task}: In this stage, we combine the parameters from the NLI stage with the pre-trained language model and fine-tune them sequentially for the multiple-choice reading comprehension task on the ViMMRC 2.0 dataset.
\end{itemize}

In addition, we apply the multi-step attention network (MAN) as a proposed top-level classifier in \cite{jin2020mmm} to calculate the attention score between the reading passage and the pair of questions and options as the answers. To represent the passages, questions, and options to a vector of tokens, we use the BERTology pre-trained language models, as mentioned in Section 4.1.
\section{Empirical Results}
\label{results}
\subsection{BERTology Models Results}
Table \ref{tab_results} illustrates the empirical results of the Bertologies model on the ViMMRC 2.0 corpus. According to \cite{9247161,9352127}, we use the Accuracy score to measure the performance of the models. From the results, the viBERT model obtains the best performance on the ViMMRC 2.0 corpus, which is 50.88\% on the dev set and 53.47\% on the test set. To indicate the efficiency of the new corpus with the previous one, we train BERTology models on the ViMMRC 2.0 dataset and let those models predict on the dev set and test set of the old version of ViMMRC \cite{9247161}. From the results in Table \ref{tab_results_ver1}, it can be shown that the performance of the models trained on the new version is better than the best baseline results in \cite{9247161,9352127}. 

\begin{table}[H]
    \centering
    \caption{Empirical results on the ViMMRC 2.0}
    \label{tab_results}
    \begin{tabular}{lcc}
        \toprule
        \textbf{Models} & \textbf{Dev (\%)} & \textbf{Test (\%)} \\
        \midrule
        mBERT & 47.51 & 47.79 \\ 
        XLM-R & 49.82 & 51.84 \\
        viBERT & \textbf{50.88} & \textbf{53.47} \\
        BERT4News & 40.42 & 41.74 \\
        \bottomrule
    \end{tabular}
\end{table}

In addition, the viBERT models show the best results on the test set and development set of the ViMMRC 2.0. In contrast, when applied to the test set and development set of the previous ViMMRC in \cite{9247161}, BERT obtains higher results than viBERT and is the highest performance model, which is 86.73\% on the dev set and 82.29\% on the test set, according to Table \ref{tab_results_ver1}. Besides, we also found that the XLM-R model is interesting for both cases. XLM-R although is not the highest result, it is a runner-up model on both ViMMRC 2.0 and the 1.0 version. 

\begin{table}[H]
    \centering
    \caption{Empirical results on the ViMMRC 1.0}
    \label{tab_results_ver1}
    \begin{tabular}{lcc}
        \toprule
        \textbf{Models} & \textbf{Dev (\%)} & \textbf{Test (\%)} \\
        \midrule
        mBERT & \textbf{86.73} & \textbf{82.29} \\
        XLM-R & 85.03 & 81.32 \\
        ViBERT & 79.59 & 70.42 \\
        BERT4News & 50.68 & 49.41 \\
        mBERT (cased) \cite{9352127} & 68.02 & 60.50 \\
        Boosted score + ELMO  \cite{9247161}& 65.99 & 61.81 \\
        \bottomrule
    \end{tabular}
\end{table}

\subsection{Multi-stage Models Results}
Table \ref{results_multi_stage} illustrates the results of the mBERT and viBERT models when applied to the MAN mechanism and the NLI task. With the MAN mechanism, the results of both viBERT and mBERT models are improved. Moreover, with the integration of the NLI task, the results of the two viBERT and mBERT models are the highest on the test set of the ViMMRC 2.0 dataset, which is 55.64\% for mBERT+MAN+NLI and 58.81\% for the viBERT+MAN+NLI. The viBERT model with the MAN mechanism and the NLI task obtains the highest results on the test set of the ViMMRC 2.0 dataset. In general, the integration of the NLI improves the performance of models for reading comprehension.

\begin{table}[H]
    \centering
    \caption{The results of multi-stage models on the ViMMRC 2.0 dataset}
    \label{results_multi_stage}
    \begin{tabular}{lcc}
        \toprule
        \textbf{Models} & \textbf{Dev (\%)} & \textbf{Test (\%)} \\
        \midrule
        mBERT & 47.51 & 47.79 \\
        mBERT + MAN & 51.60 & 52.93 \\
        mBERT + NLI & 52.13	& \textbf{56.99} \\
        mBERT + NLI + MAN & \textbf{53.37} & 55.64 \\
        viBERT & 50.88 & 53.47 \\
        viBERT + MAN & \textbf{56.03} & 57.17 \\
        viBERT + NLI & 53.55 & 56.81 \\
        viBERT + MAN + NLI & 55.32 & \textbf{58.81} \\
        \bottomrule
    \end{tabular}
\end{table}

Additionally, when we evaluate the mBERT models with the MAN and NLI tasks on the ViMMRC 1.0 \cite{9247161}, the accuracy seems to decrease, which is lower than the original mBERT without NLI and MAN. In contrast, for the viBERT model, the MAN and NLI help increase the performance of the model significantly. According to the results in Table \ref{results_multi_stage_ver1}, the accuracy of the viBERT with MAN and NLI, and mBERT models are much higher than other baselines in \cite{9247161} and \cite{9352127}. 

\begin{table}[H]
    \centering
    \caption{The results of multi-stage models on the ViMMRC 1.0 dataset}
    \label{results_multi_stage_ver1}
    \begin{tabular}{lcc}
        \toprule
        \textbf{Models} & \textbf{Dev (\%)} & \textbf{Test (\%)} \\
        \midrule
        mBERT & \textbf{86.73} & \textbf{82.29} \\
        mBERT + MAN & 76.19	& 70.23 \\
        mBERT + NLI & 78.57 &	71.79 \\
        mBERT + MAN + NLI & 70.07 & 65.95 \\
        viBERT & 79.59 & 70.42 \\
        viBERT + MAN & 80.27 & 73.93 \\
        viBERT + NLI & 80.95 & 72.96 \\
        viBERT + MAN + NLI & \textbf{84.01} & \textbf{80.16} \\
        mBERT (cased) \cite{9352127}  & 68.02 & 60.50 \\
        Boosted score + ELMO \cite{9247161}& 65.99 & 61.81 \\
        \bottomrule
    \end{tabular}
\end{table}

Finally, according to the authors in \cite{9247161}, the size of the training data affects the performance of the model. Increasing training data will improve the accuracy. From the experimental results in Table \ref{results_multi_stage_ver1} and Table \ref{tab_results_ver1}, it can be seen that the accuracy is improved when the model trained on the ViMMRC 2.0, which has more data. This also showed that ViMMRC 2.0 - an enhanced version of the previous dataset \cite{9247161}, is efficient with the multiple-choice reading comprehension models.
\section{Error Analysis and Discussion}
\label{error_analysis}
\subsection{The challenge of higher grade and the type of questions}

\begin{figure}[H]
        \centering
        \frame{\includegraphics[width=.9\textwidth]{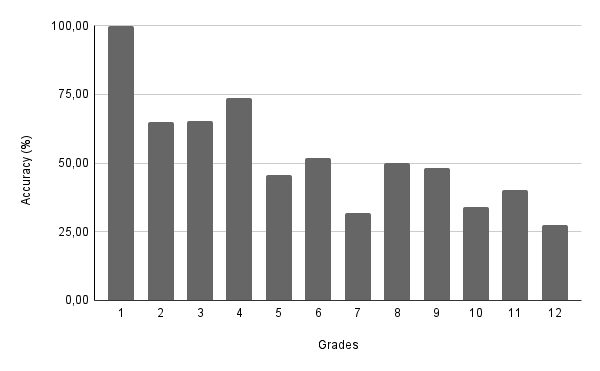}}
        \caption{Performance of viBERT model on the development set of the ViMMRC 2.0}
        \label{fig_model_by_grade}
\end{figure}

As shown in Section \ref{dataset}, the difficulty of the reading passage and the questions increase in higher grades. Hence, Table \ref{tab_result_grades} shows the accuracy of the best baseline model - the viBERT on the development set of the ViMMRC 2.0 by each grade, from the $1^{st}$ to the $12^{th}$ grades. From the results in Table \ref{tab_result_grades} and the illustration in Fig \ref{fig_model_by_grade}, the performance of the model decreases when the grade goes higher. Especially, the model obtained the accuracy with 100\% on the grade $1^{st}$. In contrast, the model has the lowest accuracy on the Grade $12^{th}$. To have an overview of the performance of the reading comprehension model on levels, we group the results into three levels, which comprise Elementary (from $1^{st}$ to $5^{th}$), Secondary (from $6^{th}$ to $9^{th}$), and High (from $10^{th}$ to $12^{th}$) according to the education level of Vietnamese students. It can be seen in Table \ref{tab_result_grades} that the accuracy of the model decreases in the higher levels where the question is more difficult. 

\begin{table}[H]
    \centering
    \caption{Performance results by each grade of the viBERT model on the dev set of the ViMMRC 2.0}
    \label{tab_result_grades}
    \resizebox{\textwidth}{!}{
    \begin{tabular}{ccccccccccccc}
        \toprule
        \textbf{Grade} & $1^{st}$ & $2^{nd}$ & $3^{rd}$ & $4^{th}$ & $5^{th}$ & $6^{th}$ & $7^{th}$ & $8^{th}$ & $9^{th}$ & $10^{th}$ & $11^{th}$ & $12^{th}$ \\
        \textbf{Accuracy (\%)} & 100 & 64.81 & 65.21 & 73.58 & 45.78 & 52.00 & 31.81 & 50.00 & 48.07 & 34.04 & 40.00 & 27.27 \\
        \cmidrule{2-6} \cmidrule(l){7-10} \cmidrule(l){11-13}
        \multirow{2}{*}{\textbf{Accuracy by level(\%)}} & \multicolumn{5}{c}{\textit{Elementary school}} & \multicolumn{4}{c}{\textit{Secondary school}} & \multicolumn{3}{c}{\textit{High school}} \\
        & \multicolumn{5}{c}{61.67} & \multicolumn{4}{c}{46.28} & \multicolumn{3}{c}{34.61} \\
        \bottomrule
    \end{tabular}
    }
\end{table}

In addition, we investigate the performance of models by types of questions as mentioned in Section \ref{dataset}. We take the viBERT and XLM-R models as a result of our experiment. The types of questions in Table \ref{tbl_result_quest} are referred from Table \ref{tab_ques_type}. It can be seen that both models have the same performance with questions in type 3 and type 4, while the performance has a small difference between models in question type 1 and type 2. Therefore, the challenge of the machine reading comprehension models focuses on type 3 and type 4 questions where both robust models have similar accuracy.

\begin{table}[H]
    \centering
    \caption{The performance of the viBERT and XLM-R by types of questions}
    \label{tbl_result_quest}
    \begin{tabular}{ccc}
        \toprule
        \textbf{Types of questions} & \textbf{viBERT Accuracy (\%)} & \textbf{XLM-R Accuracy (\%)} \\
        \midrule
        Type 1 & 49.43 & 48.53 \\
        Type 2 & 63.54 & 62.50 \\
        Type 3 & 60.00 & 60.00 \\
        Type 4 & 16.66 & 16.66 \\
        \bottomrule
    \end{tabular}
\end{table}

\subsection{The efficiency of multi-stage models}
Table \ref{tbl_result_quest_nli} illustrates the results of the viBERT model when integrating the MAN mechanism and NLI modules on four different types of questions, as shown in Table \ref{tab_ques_type}. It can be seen that the MAN with NLI has notably improved the accuracy of the Wh-questions. Besides, the appearance of NLI along with the MAN mechanism has improved the performance of models on question type 4. The MAN mechanism gives better results for question type 2 (the listed option questions), while the accuracy decreases when attaching MAN with NLI. Specifically, the NLI mechanism helps improve the performance of questions in Type 3.

\begin{table}[H]
    \centering
    \caption{The performance of the viBERT by types of questions when applying MAN and NLI mechanism}
    
    \label{tbl_result_quest_nli}
    \begin{tabular}{ccccc}
        \toprule
        \textbf{Types of questions} & \textbf{Original} & \textbf{MAN} & \textbf{NLI} & \textbf{MAN+NLI} \\
        \midrule
        Type 1 & 49.93 & 55.05 ↑ & 51.91 ↑ & \textbf{55.73} ↑ \\
        Type 2 & 63.54 & \textbf{67.70} ↑ & \textbf{66.67} ↑ & 59.37 \\
        Type 3 & 60.00 & 60.00 & \textbf{66.67} ↑ & 60.00 \\
        Type 4 & 16.66 & 16.66 & 16.67 & \textbf{22.22} ↑ \\
        \bottomrule
    \end{tabular}
\end{table}

\begin{figure}[H]
        \centering
        \frame{\includegraphics[width=.9\textwidth]{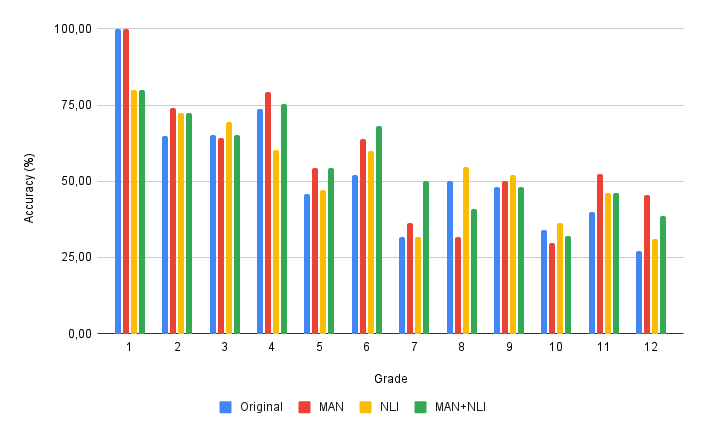}}
        \caption{Performance of viBERT model with MAN and NLI on the development set of the ViMMRC 2.0}
        \label{fig_results_by_grades}
\end{figure}

In addition, we investigate the efficiency of MAN and NLI with the viBERT model by each grade, as shown in Table \ref{tab_result_grades_nli}. Figure \ref{fig_results_by_grades} illustrates the efficiency of MAN and NLI to the viBERT models on the development set. In general, the MAN and NLI help improve the accuracy of the multiple-choice reading comprehension model on the higher grade, from the $6^{th}$ to $12^{th}$ grades. Generally, the MAN and NLI help enhance the ability of reading comprehension models for reading passages on higher grades. 

\begin{table}[H]
    \centering
    \caption{Performance results by each grade of the viBERT model with MAN and NLI on the dev set of the ViMMRC 2.0}
    \label{tab_result_grades_nli}
    \resizebox{\textwidth}{!}{
    \begin{tabular}{cccccccccccccc}
        \toprule
        \textbf{Grade} & $1^{st}$ & $2^{nd}$ & $3^{rd}$ & $4^{th}$ & $5^{th}$ & $6^{th}$ & $7^{th}$ & $8^{th}$ & $9^{th}$ & $10^{th}$ & $11^{th}$ & $12^{th}$ & \textbf{Mean} \\
        \midrule
        \textbf{Original(\%)} & 100 & 64.81 & 65.21 & 73.58 & 45.78 & 52.00 & 31.81 & 50.00 & 48.07 & 34.04 & 40.00 & 27.27 & 56.35 \\
        \textbf{MAN(\%)} & 100 & \textbf{74.07} & 64.13 & \textbf{79.24} & \textbf{54.21} & 64.00 & 36.36 & 31.81 & 50.00 & 29.78 & \textbf{52.30} & \textbf{45.45} & 56.77 \\
        \textbf{NLI(\%)} & 80.00 & 72.22 & \textbf{69.57} & 60.38 & 46.99 & 60.00 & 31.82 & \textbf{54.55} & \textbf{51.92} & \textbf{36.17} & 46.15 & 31.10 & 53.41 \\
        \textbf{MAN+NLI(\%)} & 80.00 & 72.22 & 65.21 & 75.47 & \textbf{54.21} & \textbf{68.00} & \textbf{50.00} & 40.90 & 48.07 & 31.91 & 46.15 & 38.63 & 55.89 \\
        \bottomrule
    \end{tabular}
    }
\end{table}

However, the MAN and NLI are not improving the performance of reading comprehension models in all grades. According to Figure \ref{fig_results_by_grades} and Table \ref{tab_result_grades_nli}, there are three grades, including Grade $1^{st}$, Grade $8^{th}$ and Grade $10^{th}$ that have lower accuracy when integrating the MAN and NLI mechanism. Especially on the Grade $1^{st}$, the accuracy of viBERT with MAN+NLI is dramatically decreased compared to the original viBERT. However, due to the few number of questions on grade $1^{st}$ on the development set (according to Table \ref{tab_overall_stat}), this result does not impact much on the whole performance of models. 

\subsection{Discussion}
We investigate the prediction of the viBERT models with MAN on question type 2 - the listed options question on the development set. From the observation, we see that 14 over 31 wrong answer questions have chosen the answer with the pattern "all of the choices ..." (Sample No 1 and No 2 described in Table \ref{tab_error_answers}), which accounts for nearly 45\% of wrong predictions. This bias of the model is the same as humans do when they seek the correct answer from the reading passage.

Besides, to extract the correct answers, the models have to understand the deep context of the reading passage and even the extra knowledge to extract the correct answers. For example, in Sample No 3 in Table \ref{tab_error_answers}, the content of the question is not directly shown in the reading passage. To find the correct answers, the model has to understand the grammar of Vietnamese sentences and paragraphs to define the correct punctuation as an answer to the questions. Specifically, to give the correct answer for the question in Sample No. 4., the model not only has to understand the question, which contains the order of facts and descriptive text but also has to capture the meaning of the whole reading passage to give a correct choice. Moreover, the choices in this sample have no direct description text. They are written by a list of numerals indicating the order of facts as denoted in the questions. To overcome this problem, the MRC models must have a better ability to capture the "macrostructure"\cite{sugawara-etal-2021-benchmarking} - a structure that represents the global connection between propositions instead of the local representation of facts like text-based inference (microstructure, according to \cite{sugawara-etal-2021-benchmarking}) and lexicons surfaces. Currently, our baseline model just captures the microstructure of text representation between the passages, the question, and each answer choice since its performance is still at a moderate level.

\begin{table}[H]
    \centering
    \caption{Several wrong answers sample from the development set. The content of the reading passage is described in  \ref{append_passage}}
    \label{tab_error_answers}
    \resizebox{\textwidth}{!}{
    \begin{tabular}{cp{15cm}cc}
        \toprule
        \textbf{\#} & \textbf{Questions} & \textbf{Correct} & \textbf{Prediction} \\
        \midrule
        1 & \makecell[{{p{15cm}}}]{
            \textbf{Passage}: \textit{passage\#01} \\
            \textbf{Question:} Trong ba đứa cháu, ai là đứa cháu được ông nhận xét là còn thơ dại? \emph{(In three children, who is claimed as innocent?)}\\
            A. Vân \\
            B. Việt \\
            C. Xuân \\
            D. Cả 3 bạn trên \emph{(All of those above)}\\
        } & A & D\\
        \hline
        2 & \makecell[{{p{15cm}}}]{
            \textbf{Passage}: \textit{passage\#02} \\
            \textbf{Question:} Dựa vào diễn biến sự việc và tâm trạng nhân vật, nên chia đoạn trích thành mấy đoạn nhỏ là hợp lí? \emph{(Based on the happenings of the event and the character's mood, how many parts should the excerpt be divided into?)}\\
            A. 2 đoạn (Trước cảnh thề nguyền – Cảnh thề nguyền) \emph{(2 parts (Before the oath - The oath))}\\
            B. 3 đoạn (Kiều sang nhà Kim Trọng – Kim tỉnh giấc – Thề nguyền) \emph{(3 parts (Kieu goes to Kim Trong's house - Kim wakes up - The oath))}\\
            C. 4 đoạn (cứ 5 dòng thơ một đoạn) \emph{(4 parts (1 every 5 lines))} \\
            D. A và B đều được \emph{(A and B are correct)}\\
        } & A & D \\
        \hline
        3 & \makecell[{{p{15cm}}}]{
        \textbf{Passage}: \textit{passage\#03} \\
        \textbf{Question:} Dấu câu nào không dùng để tách thành phần câu được mở rộng trong bài văn này? \emph{(What punctuation is not used to separate the additional information from sentences in this text?)}\\
        A. Dấu ngoặc đơn \emph{(brackets)}\\
        B. Dấu hai chấm \emph{(colon)}\\
        C. Dấu phẩy \emph{(comma)}\\
        D. Dấu ngoặc đơn và dấu phẩy \emph{(brackets and comma)}\\
        } & B & D \\
        \hline
        4 & \makecell[{{p{15cm}}}]{
        \textbf{Passage}: \textit{passage\#04} \\
        \textbf{Question:}: Hãy sắp xếp các chi tiết dưới đây theo đúng thứ tự xuất hiện trong truyện Sơn Tinh, Thủy Tinh. \emph{(Let arrange these facts from the Son Tinh Thuy Tinh story by order of appearance)} \\
        1. Hùng Vương thứ mười tám nêu ra yêu cầu về lễ vật. \emph{(Hùng King XVIII gives the requirements of the marriage gift)} \\
        2. Sơn Tinh đem lễ vật đến trước và cưới được vợ. \emph{(Son Tinh comes first and married the bride)} \\
        3. Vua Hùng tổ chức kén rể cho Mị Nương. \emph{(Hùng King XVIII organizes a wedding for his daughter)} \\
        4. Sơn Tinh – Thủy Tinh đánh nhau ròng rã mấy tháng trời. \emph{(Son Tinh and Thuy Tinh keep fighting for months)} \\
        A. (1) - (2) - (3) - (4). \\
        B. (1) - (3) - (2) - (4). \\
        C. (3) - (1) - (2) - (4). \\
        D. (1) - (3) - (4) - (2). \\
        } & C & D \\
        \bottomrule
    \end{tabular}
    }
\end{table}
\section{Conclusion and Future Work}
\label{conclusion}
In this paper, we provide the ViMMRC 2.0 - a new and enhanced version dataset for multiple-choice reading comprehension of Vietnamese text. The dataset was manually collected from the Vietnamese Student Textbook and is standardized by humans. This dataset contains 699 reading passages with 5,273 questions for students from $1^{st}$ to $12^{th}$ grades. In this new version, the difficulty of reading passages and questions has been increased. The reading passage contains two forms, including prose and poems, and the questions have various types that challenge the reading comprehension model in seeking the correct answers. The dataset is used to train and evaluate the ability of computers to read and comprehension of human texts. In comparison with ViMRRC 1.0, the new dataset shows a significant improvement in the results of reading comprehension models. Overall, this dataset supports the construction and evaluation of intelligent systems in understanding human language, which helps boost the research in artificial intelligence and cognitive intelligence.

Besides the ViMMRC 2.0 dataset, we also propose a multi-stage approach that integrates the multi-step attention network (MAN) with the natural language inference (NLI) task on the ViNLI dataset \cite{huynh-etal-2022-vinli} to improve the performance of BERTology models for reading comprehension. From the empirical results, the viBERT model with MAN and NLI obtains the best results on both ViMMRC 2.0 and the previous version \cite{9247161}. However, although the BERTology model and the MAN with NLI have the best performance on the current ViMMRC 2.0 dataset, the accuracy of the reading comprehension models is still in the medium range. We hope our new dataset will motivate the next research to improve the ability of computers to understand Vietnamese texts. 

Last, from the error analysis on questions, we found that the challenge is about the implicit meaning of the questions and the content of options. Hence, our next research will focus on constructing a situation model that can have the ability to represent the macrostructure levels of propositions \cite{sugawara-etal-2021-benchmarking}. Besides, our next study is to extend the MRC model to not only text but also image, sound, and background data. Apart from the effort to strengthen the ability of the MRC model and enlarge the MRC dataset with more challenging aspects to test the MRC models, the human-related features such as emotional state and practical capability also affect the performance of machine learning model \cite{wu2022survey}. 

\section*{Acknowledgements}
This research was supported by The VNUHCM-University of Information Technology's Scientific Research Support Fund

\appendix
\section{Reading passages}
\label{append_passage}
\textbf{Passage number: passage\#01} \\
 1. Sau một chuyến đi xa, người ông mang về bốn quả đào. Ông bảo vợ và các cháu: \\
            - Quả to này xin phần bà. Ba quả nhỏ hơn phần các cháu.
            Bữa cơm chiều hôm ấy, ông hỏi các cháu: \\
            - Thế nào, các cháu thấy đào có ngon không? \\
            2. Cậu bé Xuân nói: \\
            - Đào có vị rất ngon và mùi thật là thơm. Cháu đã đem hạt trồng vào một cái vò. Chẳng bao lâu, nó sẽ mọc thành một cây đào to đấy, ông nhỉ? \\
            - Mai sau cháu sẽ làm vườn giỏi. \\
            - Ông hài lòng nhận xét. \\
            3. Cô bé Vân nói với vẻ tiếc rẻ: \\
            - Đào ngon quá, cháu ăn hết mà vẫn thèm. Còn hạt thì cháu vứt đi rồi. \\
            - Ôi cháu của ông còn thơ dại quá! \\
            4. Thấy Việt chỉ chăm chú nhìn vào tấm khăn trải bàn, ông ngạc nhiên hỏi: \\
            - Còn Việt, sao cháu chẳng nói gì thế? \\
            - Cháu ấy ạ? Cháu mang đào cho Sơn. Bạn ấy bị ốm. Nhưng bạn ấy không muốn nhận. Cháu đặt quả đào lên trên giường rồi trốn về. \\
            - Cháu là người có tấm lòng nhân hậu!  \\
            - Ông lão thốt lên và xoa đầu đứa cháu nhỏ. \\
\\
\textbf{Passage number: passage\#02} \\
        Cửa ngoài vội rủ rèm the,\\
        Xăm xăm băng lối vườn khuya một mình.\\
        Nhặt thưa gương giọi đầu cành,\\
        Ngọn đèn trông lọt trướng huỳnh hắt hiu.\\
        Sinh vừa tựa án thiu thiu,\\
        Dở chiều như tỉnh dở chiều như mê.\\
        Tiếng sen sẽ động giấc hè,\\
        Bóng trăng đã xế hoa lê lại gần.\\ \\
        Bâng khuâng đỉnh Giáp non thần.\\
        Còn ngờ giấc mộng đêm xuân mơ màng.\\
        Nàng rằng: “Khoảng vắng đêm trường,\\
        Vì hoa nên phải trổ đường tìm hoa.\\
        Bây giờ rõ mặt đôi ta,\\
        Biết đâu rồi nữa chẳng là chiêm bao?”\\
        Vội mừng làm lễ rước vào,\\
        Đài sen nối sáp lò đào thêm hương.\\
        Tiên thề cùng thảo một chương,\\
        Tóc mây một món dao vàng chia đôi.\\
        Vừng trăng vằng vặc giữa trời,\\
        Đinh ninh hai miệng một lời song song.\\
        Tóc tơ căn vặn tấc lòng,\\
        Trăm năm tạc một chữ đồng đến xương. \\
\\
\textbf{Passage number: passage\#03}\\
        Người Việt Nam ngày nay có lí do đầy đủ và vững chắc để tự hào với tiếng nói của mình. Và để tin tưởng hơn nữa vào tương lai của nó.\\
        Tiếng Việt có những đặc sắc của một thứ tiếng đẹp, một thứ tiếng hay. Nói thế có nghĩa là nói rằng: tiếng Việt là một thứ tiếng hài hoà về mặt âm hưởng, thanh điệu mà cũng rất tế nhị, uyển chuyển trong cách đặt câu. Nói thế cũng có nghĩa là nói rằng: tiếng Việt có đầy đủ khả năng để diễn đạt tình cảm, tư tưởng của người Việt Nam và để thoả mãn cho yêu cầu của đời sống văn hoá nước nhà qua các thời kì lịch sử.\\
        Tiếng Việt, trong cấu tạo của nó, thật sự có những đặc sắc của một thứ tiếng khá đẹp. Nhiều người ngoại quốc sang thăm nước ta và có dịp nghe tiếng nói của quần chúng nhân dân ta, đã có thể nhận xét rằng: tiếng Việt là một thứ tiếng giàu chất nhạc. Họ không hiểu tiếng ta, và đó là một ấn tượng, ấn tượng của người “nghe” và chỉ nghe thôi. Tuy vậy lời bình phẩm của họ có phần chắc không phải chỉ là một lời khen xã giao. Những nhân chứng có đủ thẩm quyền hơn về mặt này cũng không hiếm. Một giáo sĩ nước ngoài (chúng ta biết rằng nhiều nhà truyền đạo Thiên Chúa nước ngoài cũng là những người rất thạo tiếng Việt), đã có thể nói đến tiếng Việt như là một thứ tiếng “đẹp” và “rất rành mạch trong lối nói, rất uyển chuyển trong câu kéo, rất ngon lành trong những câu tục ngữ”. Tiếng Việt chúng ta gồm có một hệ thống nguyên âm và phụ âm khá phong phú. Tiếng ta lại giàu về thanh điệu. Giọng nói của người Việt Nam, ngoài hai thanh bằng (âm bình và dương bình) còn có bốn thanh trắc. Do đó, tiếng Việt có thể kể vào những thứ tiếng giàu hình tượng ngữ âm như những âm giai trong bản nhạc trầm bổng. Giá trị của một tiếng nói cố nhiên không phải chỉ là câu chuyện chất nhạc. Là một phương tiện trao đổi tình cảm ý nghĩ giữa người với người, một thứ tiếng hay trước hết phải thoả mãn được nhu cầu ấy của xã hội. Về phương diện này, tiếng Việt có những khả năng dồi dào về phần cấu tạo từ ngữ cũng như về hình thức diễn đạt. Từ vựng tiếng Việt qua các thời kì diễn biến của nó tăng lên mỗi ngày một nhiều. Ngữ pháp cũng dần dần trở nên uyển chuyển hơn, chính xác hơn. Dựa vào đặc tính ngữ âm của bản thân mình, tiếng Việt đã không ngừng đặt ra những từ mới, những cách nói mới hoặc Việt hoá những từ và những cách nói của các dân tộc anh em và các dân tộc láng giềng, để biểu hiện những khái niệm mới, để thoả mãn yêu cầu của đời sống văn hoá ngày một phức tạp về mọi mặt kinh tế, chính trị, khoa học, kĩ thuật, văn nghệ, ... \\
        Chúng ta có thể khẳng định rằng: cấu tạo của tiếng Việt, với khả năng thích ứng với hoàn cảnh lịch sử như chúng ta vừa nói trên đây, là một chứng cớ khá rõ về sức sống của nó. \\
\\
\textbf{Passage numner: passage\#04}\\
        Hùng Vương thứ mười tám có một người con gái tên là Mị Nương, người đẹp như hoa, tính nết hiền dịu. Vua cha yêu thương nàng hết mực, muốn kén cho con một người chồng thật xứng đáng.\\
        Một hôm có hai chàng trai đến cầu hôn"[1]". Một người ở vùng núi Tản Viên"[2]" có tài lạ: vẫy tay về phía đông, phía đông nổi cồn bãi; vẫy tay về phía tây, phía tây mọc lên từng dãy núi đồi. Người ta gọi chàng là Sơn Tinh. Một người ở miền biển, tài năng cũng không kém: gọi gió, gió đến; hô mưa, mưa về. Người ta gọi chàng là Thuỷ Tinh. Một người là chúa vùng non cao, một người là chúa vùng nước thẳm, cả hai đều xứng đáng làm rể vua Hùng. Vua Hùng băn khoăn không biết nhận lời ai, bèn cho mời các Lạc hầu"[3]" vào bàn bạc. Xong, vua phán"[4]":\\
        - Hai chàng đều vừa ý ta, nhưng ta chỉ có một người con gái, biết gả cho người nào? Thôi thì ngày mai, ai đem sính lễ"[5]" đến trước, ta sẽ cho cưới con gái ta.\\
        Hai chàng tâu"[6]" hỏi đồ sính lễ cần sắm những gì, vua bảo: “Một trăm ván cơm nếp, một trăm nệp bánh chưng và voi chín ngà, gà chín cựa, ngựa chín hồng mao"[7]", mỗi thứ một đôi”.\\
        Hôm sau, mới tờ mờ sáng, Sơn Tinh đã đem ra đầy đủ lễ vật đến rước Mị Nương về núi.\\
        Thuỷ Tinh đến sau, không lấy được vợ, đùng đùng nổi giận, đem quân đuổi theo đòi cướp Mị Nương. Thần hô mưa, gọi gió làm thành dông bão rung chuyển cả đất trời, dâng nước sông lên cuồn cuộn đánh Sơn Tinh. Nước ngập ruộng đồng, nước ngập nhà cửa, nước dâng lên lưng đồi, sườn núi, thành Phong Châu như nổi lềnh bềnh trên một biển nước.\\
        Sơn Tinh không hề nao núng"[8]". Thần dùng phép lạ bốc từng quả đồi, đời từng dãy núi, dựng thành luỹ đất, ngăn chặn dòng nước lũ. Nước sông dâng lên bao nhiêu, đồi núi cao lên bấy nhiêu. Hai bên đánh nhau ròng rã mấy tháng trời, cuối cùng Sơn Tinh vẫn vững vàng mà sức Thuỷ Tinh đã kiệt. Thần Nước đành rút quân.\\
        Từ đó, oán nặng, thù sâu, hằng năm Thuỷ Tinh làm mưa làm gió, bão lụt dâng nước đánh Sơn Tinh. Nhưng năm nào cũng vậy, Thần Nước đánh mỏi mệt, chán chê vẫn không thắng nổi Thần Núi để cướp Mị Nương, đành rút quân về.\\\\
        "[1]" Xin được lấy làm vợ (cầu: tìm, kiếm, xin; hôn: lấy vợ, lấy chồng).\\
        "[2]" Núi cao ở huyện Ba Vì, tỉnh Hà Tây, cũng gọi là núi Ba Vì. Núi có ba đỉnh, đỉnh cao nhất 1281 mét, ngọn giữa có hình thắt cổ bồng, trên toả ra như cái tán nên gọi là Tản Viên. Thần núi Tản Viên (Sơn Tinh) được coi là vị thần linh thiêng nhất của nước ta xưa.\\
        "[3]" Chức danh của các vị quan cao nhất giúp vua Hùng trông coi việc nước.\\
        "[4]" Truyền bảo (từ được dùng khi người truyền bảo là vua chúa, thần linh, cũng có thể là người bề trên nói chung).\\
        "[5]" Là vật nhà trai đem đến nhà gái để xin cưới.\\
        "[6]" Thưa trình (từ dùng khi quan, dân nói với vua chúa, thần linh).\\
        "[7]" Ở đây chỉ bờm ngựa màu hồng.\\
        "[8]" Lung lay, không vững lòng tin ở mình nữa. \\

\section{Data availability and source code}
This dataset is freely available for research purposes only. Users need to sign the data agreement form before downloading it. Please contact the authors for data inquiries. 

\textbf{Source code for the baseline:} \url{https://github.com/sonlam1102/vimmrc2}

\textbf{Source code for the annotation tool:} \url{https://github.com/sonlam1102/vimmrc2.0-annotation-tools}

\bibliographystyle{ws-ijalp}
\bibliography{references}

\end{document}